\documentclass[10pt,journal,compsoc]{IEEEtran}
\usepackage[nocompress]{cite}
\usepackage{bm}
\usepackage{amsfonts}
\usepackage{amssymb}

\usepackage{multirow}
\usepackage{array}
\usepackage{booktabs}

\usepackage{amsmath}
\usepackage[ruled,vlined]{algorithm2e}
\usepackage{algorithmic}
\newtheorem{problem}{Problem}

\usepackage[pdftex]{graphicx}
\ifCLASSOPTIONcompsoc
  \usepackage[caption=false,font=footnotesize,labelfont=sf,textfont=sf]{subfig}
\else
  \usepackage[caption=false,font=footnotesize]{subfig}
\fi
\begin{document}

\title{Neighboring Backdoor Attacks on Graph Convolutional Network}

\author{
    Liang~Chen,~
    Qibiao~Peng,~\
    Jintang~Li,~\
    Yang~Liu,~\
    Jiawei~Chen,~\
    Yong~Li,~\
    and~Zibin~Zheng
}

\markboth{Journal of \LaTeX\ Class Files,~Vol.~14, No.~8, August~2015}%
{Shell \MakeLowercase{\textit{et al.}}: Bare Demo of IEEEtran.cls for Computer Society Journals}



\IEEEtitleabstractindextext{%
\begin{abstract}
Backdoor attacks have been widely studied to hide the misclassification rules in the normal models, which are only activated when the model is aware of the specific inputs (i.e., the trigger). However, despite their success in the conventional Euclidean space, there are few studies of backdoor attacks on graph structured data.
In this paper, we propose a new type of backdoor which is specific to graph data, called \textit{neighboring backdoor}. 
Considering the discreteness of graph data, how to effectively design the triggers while retaining the model accuracy on the original task is the major challenge. To address such a challenge, we set the trigger as a single node, and the backdoor is activated when the trigger node is connected to the target node. To preserve the model accuracy, the model parameters are not allowed to be modified. Thus, when the trigger node is not connected, the model performs normally. Under these settings, in this work, we focus on generating the features of the trigger node. Two types of backdoors are proposed: (1) Linear Graph Convolution Backdoor which finds an approximation solution for the feature generation (can be viewed as an integer programming problem) by looking at the linear part of GCNs. (2) Variants of existing graph attacks. We extend current gradient-based attack methods to our backdoor attack scenario. Extensive experiments on two social networks and two citation networks datasets demonstrate that all proposed backdoors can achieve an almost 100\% attack success rate while having no impact on predictive accuracy. 
\end{abstract}

% TODO: change to IEEE keywords
\begin{IEEEkeywords}
Graph neural networks, backdoor attacks
\end{IEEEkeywords}}

\maketitle

\section{Introduction}
\begin{figure}[t]
\centering
\includegraphics[width=0.48\textwidth]{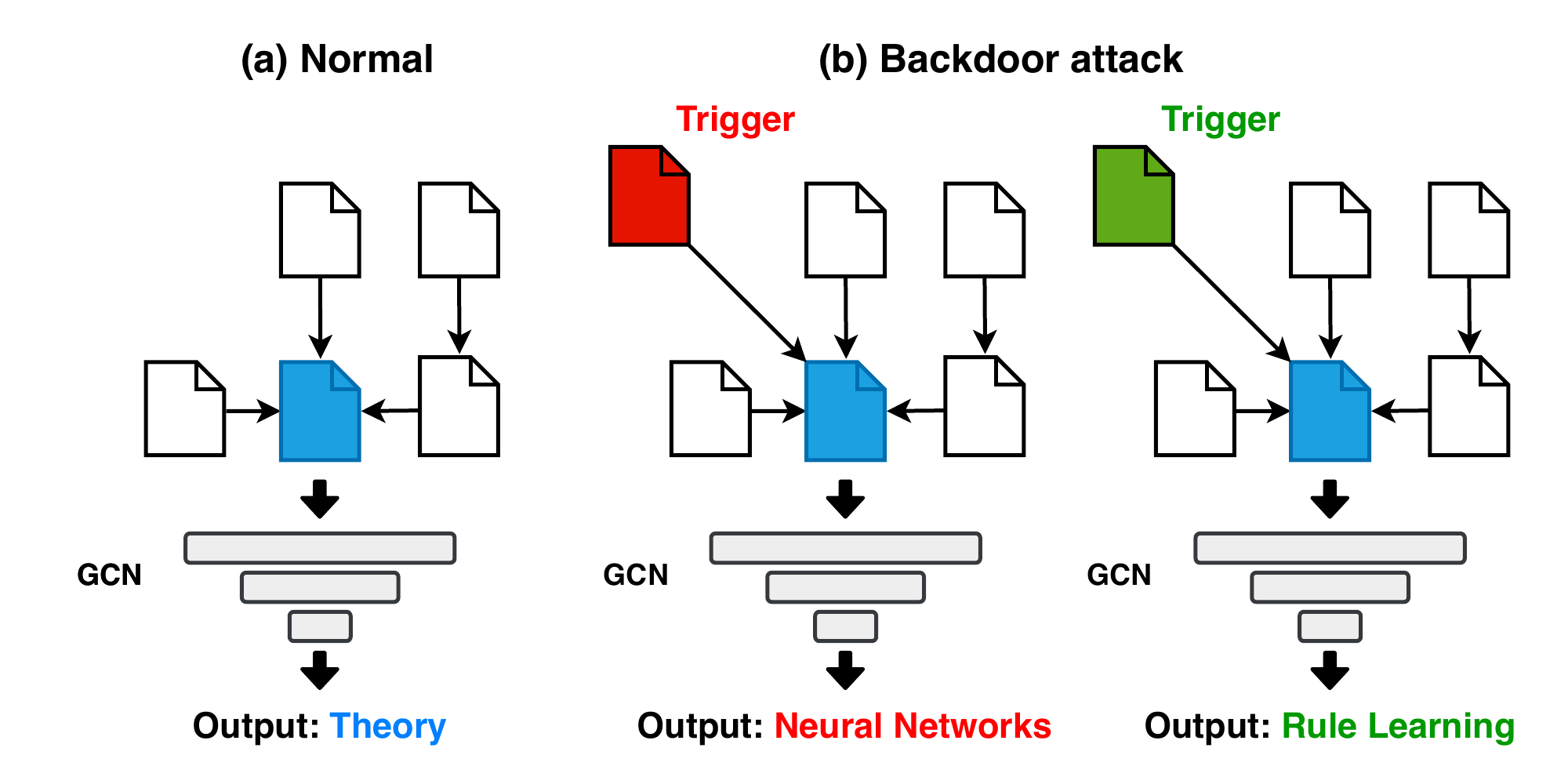}
\caption{Illustration of GCN backdoor attacks on Cora citation network. (a) GCN works normally without triggers; (b) adding different triggers can mislead the model to predict different target labels.} \label{fig:illustration}
\end{figure}

Due to the strong ability of learning graph structure data, graph convolutional networks (GCNs)~\cite{DBLP:conf/iclr/KipfW17} have become increasingly popular in many tasks including recommender systems~\cite{DBLP:conf/sigir/Wang0WFC19,DBLP:conf/sigir/0001DWLZ020}, spammer detection~\cite{DBLP:conf/aaai/WuLXWC20}, and rumor detection~\cite{DBLP:conf/aaai/BianXXZHRH20,DBLP:conf/cikm/DongZHSL19}. To reduce the financial cost, users may train GCN on third-party platforms~\cite{DBLP:journals/corr/abs-2007-08745}, such as Azure~\footnote{https://azure.microsoft.com/} and BigML~\footnote{https://bigml.com/}; users may even directly utilize third-party pre-trained models. This provides an opportunity for malicious developers or hackers to intentionally insert backdoors into GCNs. That is, the model performs normally on original tasks while behaves incorrectly when backdoors are activated by special triggers. Figure~\ref{fig:illustration} is an example of GCN backdoor attacks on Cora citation network. As can be seen, attackers can easily change model predictions to target labels via particular triggers. A common solution~\cite{DBLP:journals/corr/abs-2007-08745,DBLP:conf/kdd/TangDLYH20} to insert backdoors and generate triggers is firstly preparing a poisoned dataset and fine-tuning the target model with the contaminated data.
%the workflow of backdoor attacks

Recently, the robustness of GCNs has attracted considerable attention and there has been a surge of researches~\cite{DBLP:conf/kdd/ZugnerAG18,DBLP:conf/icml/DaiLTHWZS18,DBLP:conf/iclr/ZugnerG19,DBLP:conf/ijcai/Wu0TDLZ19,DBLP:conf/ijcai/XuC0CWHL19,DBLP:conf/aaai/ChangRXHZC0H20,DBLP:conf/www/LiZHRCH20,DBLP:conf/www/SunWTHH20} studying how to attack GCNs on tasks including community detection, link prediction, and so on. This paper study backdoor attacks on GCNs under the context of the node classification. That is, given a graph and a few labels of nodes, predict the labels of the remaining nodes. The key idea of these attack methods is to manipulate the input graph (e.g., modifying node features, inserting/deleting edges or nodes), which misclassify inputs to wrong labels. Despite the effectiveness of these methods, they are not or can not be generalized to backdoor attacks. The major reasons are that they usually generate a unique perturbation for each node or an entire graph (i.e., no trigger) and the attacked labels are non-targeted (i.e., the attack is successful if the predicted label after the attack is different from the original label.) while backdoor attacks require the prediction change to an attacker-specified target label. Recently, a few studies also explored backdoor attacks in the graph domain~\cite{DBLP:journals/corr/abs-2006-11165, DBLP:journals/corr/abs-2006-11890}. However, these methods follow the idea of common backdoors that preparing a set of poisoned graphs with triggers and fun-tuning the GNNs, which leads to several limitations: (1) It needs retraining procedure which is computational inefficient and the model parameters would be modified thus can't be directly adopted for pre-trained GCNs. (2) Since the backdoors are injected by modifying model parameters, the performance on the clean graph may be affected even when the trigger is not activated. 

We try to bridge this gap in this paper and investigate whether such backdoors are possible for GCNs. \textit{Can we easily backdoor graph convolutional networks?} \textit{How vulnerable are graph convolutional networks to backdoor attacks?} Note that GCNs learn the representation of a node by aggregating the representations of its neighbors. In other words, nodes in a graph are highly dependent on their neighbors, which enlighten us a new way to inject backdoor to GCNs by neighborhood poisoning without retraining or modifying model parameters, which we called \textit{neighboring backdoor}. However, there remain the following challenges: (1) \textbf{Accuracy.} Previous works have demonstrated that the extra retraining procedure can potentially degrade model performance when injecting backdoors into multiple target labels~\cite{DBLP:conf/kdd/TangDLYH20}. To better sell or spread the model, it is crucial for attackers to retain the model accuracy on the original task. How to insert backdoors while do not harm the model accuracy? (2) \textbf{Efficiency.} When the graph is larger, the traditional solution, retraining a GCN on a contaminated dataset, is usually computationally expensive and time-consuming. Moreover, unlike images which consist of continuous features, the nodes' features (e.g., bag-of-words) are often discrete. How to design an efficient trigger generation algorithm in a discrete domain?

To maintain the model accuracy in the original task, the model parameters are not allowed to be modified (e.g., retraining). Instead, we focus on designing the trigger and the model prediction only changed when the trigger is connected to the target node. Therefore, when the trigger node does not appear, our methods have no influence on the model performance, which make attacks imperceptible as well. Moreover, injecting the backdoors into multiple classes simultaneously can be achieved by generating triggers that corresponding to different classes. To enhance the efficiency, the trigger is set as a single node. Thus designing the trigger is the same as generating the node features.  Solving such a problem is easier and more efficient compared to jointly consider the graph structure and node features (if the trigger is multiple nodes such as a sub-graph). An additional benefit is that employing a single node has a low cost. Under these settings, two types of backdoors are proposed in this work:

% the trigger is set as a node, which reduces the risk of being suspected and has a lower cost compared to a sub-graph (i.e., more than one node). Instead of retraining the model, we focus on designing an effective approach to generate the features of triggers. The backdoor is activated when the trigger node is connected to the target node. The advantages are that: (1) such an attack do not require retraining the target model on a contaminated dataset; (2) injecting backdoor has no influence on GCNs performance on original tasks (i.e., without trigger nodes) and makes attacks imperceptible; (3) It can inject the backdoors into multiple classes simultaneously while not degrade model performance. Based on the accessibility of data, we propose the following two types of backdoors:
\begin{itemize}
    \item {\bf Linear graph convolution backdoor (LGCB).} First, the feature generation is formally defined as an integer programming problem. To efficiently find a solution, a linear approximation is proposed whose core idea is to look at the linear part of GCNs. That is, the ReLU activation functions are removed. Compared to the following approaches, LGCB is faster while achieving compatible attack performance, and require less knowledge (we show that the graph structure is not necessary to feature generation)
    \item {\bf Gradient-based backdoors (GB).} Most current methods aim to derive a local attack through gradient information. That is, each node has its unique attack (e.g., modified different features for different nodes). However, our attack scenario requires a global trigger (i.e., different nodes connect to the same trigger). Since the key of these methods is how to utilize the gradient when the data is discrete/binary, we show that they can be extended to find a global trigger.
\end{itemize}

Our contributions are summarized as follows:
\begin{itemize}
    % \item We highlight the robustness of graph convolutional networks against the backdoor attack and show that such attacks can be achieved via a single node and modifying the model parameters is not necessary.
    \item We propose neighboring backdoor on graph convolutional network by utilizing its own connective vulnerabilities, so that modifying the model parameters is not necessary, and show that only poisoning a single neighbor is sufficient to construct the neighboring backdoor.
    \item We propose a fast and effective trigger generation model whose core idea is to approximate the optimal trigger by looking at the linear part of GCNs and extend three existing graph attacks to backdoor graph convolutional networks.
    \item Experimental results demonstrate that the proposed methods achieve all-label attacks with almost 100\% attack success rates using a single trigger node, while keeping the trigger node unnoticeable as well.
\end{itemize}

\section{Preliminaries}

\subsection{Backdoor Attack}
Backdoor attacks against machine learning systems aim to inject malicious behavior into parameterized models, which is activated when trigger patterns are presented in the model's inputs. Generally, the backdoor attack process can be summarized as a three-stage framework: trigger generation, trigger activation, and model retraining. To better understand the particularity of the backdoor on node classification tasks, we compare it with the backdoor on image classification.

The trigger generation in images is to generate a pair that consists of a specific pattern (e.g., sticker, dots, and tattoo) and an attacker-specified label. Such patterns are rarely appeared in normal input data and usually just designed as simple as possible (e.g., just a black patch) to avoid performance degrading on normal inputs. While the trigger in node classification is much different: (1) it need carefully design to be able to successfully attack by a one-node trigger, and (2) the generated trigger node may appear as a normal node so it's more unnoticeable. 

The trigger activation means embedding the trigger into inputs for the preparation of retraining or testing. In images, the trigger is added to many examples in dataset $\mathcal{D}$ to construct enough poisoned examples to learn the patterns of the trigger. While our trigger node is activated by simply connecting to the target node, and don't modify other nodes in the graph for more poisoned examples. 
% Suppose $u=n+1$ is the identity of trigger node, $\bm{x}^u$ denote the trigger, given a target node $v$, the trigger activation is formulated as:

% \begin{equation}\label{eq:3}
%     A_{ij}^\prime=
%     \left\{
%     \begin{array}{cl}
%         A_{ij}, & if\ i\leqslant n\ and\ j\leqslant\ n,\\
%         1, & if\ (i,j)=(u,v)\ or\ (v,u), \\
%         0, & otherwise.
%     \end{array}
%     \right. 
%     \bm{X}^\prime=\left[\begin{array}{c} \bm{X} \\ \bm{x}^u \end{array}\right]
% \end{equation}

% where $A_{ij}$ is origin adjacent matrix entry, $\bm{A}^{\prime}\in\mathbb{R}^{(n+1)\times (n+1)}$ is the adjacent matrix after trigger node being activated in graph, $\bm{X}^{\prime}\in\mathbb{R}^{(n+1)\times d}$ is the feature matrix augmented with trigger. For simplicity of discussion, in this paper, we focus on the case that only activates one trigger, while it's possible to activate multiple triggers simultaneously in a graph in practice.

% The backdoor attack in images usually need retrain model to associate trigger patterns and target label. In other words, patterns are poisoned to model parameters. In node classification, nodes in a graph are highly dependent on their neighbors, which enlighten us a new way to inject backdoor to GCNs by neighborhood poisoning without retraining or modifying model parameters. 
%Figure~\ref{fig:comp} shows the differences between image and graph backdoors.

% backdoor goal

\subsection{Graph Convolutional Networks}\label{sec:pre}

First, we introduce graph convolutional networks under the context of the node classification task. For notations, we use bold uppercase letters to denote matrices (e.g., $\bm{W}$), bold lowercase letters to denote vectors (e.g., $\bm{w}$), bold lowercase letters with superscript to denote row vectors (e.g., $\bm{x}^v$), with subscript to denote column vectors (e.g., $\bm{w}_c$), and non-bold letters to denote scalars or indices (e.g., $w$). The uppercase calligraphic symbols (e.g., $\mathcal{W}$) stand for sets. Our goal is to predict the associated label for each node. GCNs follow a recursive neighborhood aggregation scheme, where at each iteration of aggregation, the representation of a node is generated by aggregating the representation of its neighbors, followed by a linear transformation and non-linear activation function. After $L$ iterations of aggregation, a node's representation can capture the structural information of its $L$-hop neighbors. Given an undirected attributed graph $\mathcal{G} = (\bm{A}, \bm{X})$ that has $n$ nodes, with $\bm{A}\in \{0,1\}^{n\times n}$ denoting the adjacency matrix and $\bm{X}\in \{0,1\}^{n\times d}$ representing the nodes' binary feature where $d$ is the feature dimension, the $l$-th hidden layer of GCN is defined as:
\begin{equation}\label{eq:gcn}
\bm{H}^{(l)} = \sigma(\bm{S}\bm{H}^{(l-1)}\bm{W}^{(l-1)}) \ ,
\end{equation}
where $\bm{S} = \tilde{\bm{D}}^{-\frac{1}{2}}\bm{\tilde{A}}\tilde{\bm{D}}^{-\frac{1}{2}}$ and $\bm{\tilde{A}} = \bm{A}+\bm{I}$. $\bm{I}\in \mathbb{R}^{n\times n}$ is the identity matrix. $\tilde{\bm{D}}$ represents the degree matrix whose $i$-th entry $d_{ii}$ is $\sum_j\tilde{a}_{ij}$. $\bm{W}^{(l)}$ is the trainable weight matrix of $l$-th layer and $\sigma(\cdot)$ denotes the activation function, which we set as the ReLU. Initially, $\bm{H}^{(0)} = \bm{X}$. To perform node classification, the model prediction is normalized by a softmax function: $\text{softmax}(\bm{x})=\text{exp}(\bm{x})/\sum_{i=1}^{k}\text{exp}(x_{i})$. Consider an $L$-layer GCN, the final  prediction is derived as $Z=f(\bm{A}, \bm{X}) = \text{softmax}(\bm{H}^{(L)}) \in \mathbb{R}^{n\times k}$, where $k$ denotes the number of label classes. The GCN is learned by minimizing the cross-entropy loss on given training label node set.

% $\mathcal{V}$:

% \begin{equation}\label{eq:loss}
% % \mathcal{L}(\bm{A},\bm{X})=-\sum_{v\in\mathcal{V}} \bm{Y}_{v}\ln{\bm{Z}_{v}} + (1 - \bm{Y}_{v})\ln{(1 - \bm{Z}_{v})} \ ,
% \mathcal{L}(\bm{A},\bm{X})=-\sum_{v\in\mathcal{V}} \bm{Y}_{v}\ln{\bm{Z}_{v}} \ , 
% \end{equation}

% where $\bm{Y}_{v}\in\mathbb{R}^{k}$ is the one-hot vector for labeled node $v$.

% \begin{figure}[t]
% \centering
% \includegraphics[width=0.48\textwidth]{comparison.pdf}
% \caption{Differences between image and graph backdoors.} \label{fig:comp}
% \end{figure}

\subsection{Problem Statement}
We first introduce the concepts of trigger and backdoor in graph settings. The trigger in a graph is often generated by node injection \cite{DBLP:journals/corr/abs-2006-11165, DBLP:journals/corr/abs-2006-11890}. Generally, suppose $s$ nodes are injected to the clean graph $\mathcal{G} = (\bm{A}, \bm{X})$ and generate a backdoored graph $\mathcal{G}^{\prime} = (\bm{A}^{\prime}, \bm{X}^{\prime})$, which is formulated as:
\begin{equation}\label{eq:multi-trigger}
    \bm{A}^\prime=\left[\begin{array}{cc} \bm{A} & \bm{C}^{T} \\ \bm{C} & \bm{B} \end{array}\right] , \ 
    \bm{X}^\prime=\left[\begin{array}{c} \bm{X} \\ \bm{F} \end{array}\right]
\end{equation}
where $\bm{B}\in\{0,1\}^{s\times s}$ denotes the topological connectivities within the trigger nodes, $\bm{C}\in\{0,1\}^{s\times n}$ denotes the connectivities between the trigger nodes and the normal nodes, $\bm{F}\in\mathbb{R}^{s\times d}$ is the node features of the trigger nodes. The injected nodes and connectivities identify the trigger $\mathcal{T}=(\bm{B},\bm{C},\bm{F})$, and the trigger is called \textit{activated} in the graph $\mathcal{G}^{\prime}$ and not activated in original graph $\mathcal{G}$. 

\subsubsection{Attackers' goal}
In backdoor attacks, the attackers have two goals: (1) modify the outputs of GCNs to targeted predictions (i.e., attackers specified labels) when the trigger activated, and (2) keep the original predictions of GCNs when the trigger is not activated. Thus the attackers can launch or hide backdoor attacks on demand by toggling the activation states of the trigger.

% Because the attackers are not aiming to destroy the model, but intentionally and stealthily inject malicious function into the target model. It can be easily detected and further defended if the triggers cause noticeable wrong predictions on original tasks. In summary, attackers' goals including high attack success rate, low performance impact on normal inputs, and high backdoor stealthiness.

\subsubsection{Attackers' capability}
Firstly, to activate the trigger, the attackers need the ability to inject extra nodes and edges to the input graph. It can be achieved by user-level actions if the trigger size $s$ is small enough, for example, uploading some papers in the citation network.
Secondly, we have two attack scenarios, one is that attackers can access the parameters $\mathcal{W} = \{\bm{W}^{(0)}, \dots, \bm{W}^{(L-1)}\}$ of the model and training data $\mathcal{D}$, the other is that can only access $\mathcal{W}$. In both scenarios, attackers can't retrain the model (i.e., can read but not modify $\mathcal{W}$).

% test-time attack
% no impact on original graph
% no need to retrain the model
% however, 

\subsubsection{Threat model}

In summary, the attackers are supposed to design the trigger $\mathcal{T}$ to achieve their goals with the capability and constraints. Various threat models can be discussed based on different implementation details in this framework. First, according to different trigger size, the trigger can have multiple nodes or only a single node (i.e., $s>1$ or $s=1$). Second, the trigger patterns can be designed as topological patterns (i.e., design $\bm{B}$ and $\bm{C}$), feature patterns (i.e. design $\bm{F}$), or hybrid patterns.

Specifically, in this work, 
% we focus on the design of features and degrade the trigger's sub-graph structure into one node, which can alleviate the problem of complex combinational optimization, and the trigger can be simply injected by a single undirected edge with the target node.
we focus on the single node trigger for three reasons: 
(1) \textbf{Practical}: less trigger nodes indicate less actions the attackers need to launch the attack. Especially when the trigger size is 1, which is attractive for the attackers to be able to attack the target node by a single node injection.
(2) \textbf{Unnoticeable}: less trigger nodes indicate less structural impact on the graph.  
(3) \textbf{Efficient}: less trigger nodes indicate less parameter space.
Then Eq.~\ref{eq:multi-trigger} for single node trigger is simplified as:

\begin{equation}
    \bm{A}^\prime=\left[\begin{array}{cc} \bm{A} & \bm{c}^{T} \\ \bm{c} & 1 \end{array}\right] , \ 
    \bm{X}^\prime=\left[\begin{array}{c} \bm{X} \\ \bm{x}^{u} \end{array}\right]
\end{equation}
where $u=n+1$ is the identity of trigger node, $\bm{x}^u$ denotes the trigger features, $\bm{c}$ denotes the connectivities of the trigger node. The trigger node is only connected to the target node $v$ by default, thus $c_i=1$ for $i=v$ and $c_i=0$ for all $i\neq v$. The potential impact of connectivities will be discussed in the experiments.

\subsubsection{Problem definition}

Let $f:\mathbb{R}^{d}\to\mathbb{R}^{k}$ denote the GCN with $d$ input features and $k$ output classes and $\bm{x}^u$ represents the trigger node whose target label is $t$. Our goal is to generate the trigger features based on model $f$ so that the predicted labels of all nodes change to $t$ once they are connected to $\bm{x}^u$. Initially, all features of $\bm{x}^u$ are zero. Suppose $q\in\mathbb{N}$ denotes the budget (i.e., the number of changed features) and $\mathcal{P}_q(\bm{x}^u)$ denotes all possible trigger $\bm{x}^u$, the problem can be defined as follows:

\begin{problem}\label{pro1}
Given a trained model $f$, a target label $t$, and a budget $q$, let $\bm{h}^{v}\in\mathbb{R}^k$ denotes the model prediction for any specific node $v$ in $\mathcal{G}$. The optimal margin $(h^{v}_{c} - h^{v}_t)$ between class $t$ and class $c$ of all possible triggers $\mathcal{P}_q(\bm{x}^u)$ is:
\begin{equation}\label{eq:pro}
m_{t, c}^{v} = \min_{\bm{x}^u\in \mathcal{P}_q(\bm{x}^u)} h^{v}_{c} - h^{v}_t
\end{equation}
If $m_{t, c}^{v} < 0$ for all $c\neq t$, the node $v$ is successfully attacked by trigger $\bm{x}^u$.
\end{problem}

Since we focus on the relative value of each class, it is unnecessary to employ softmax function on model prediction. Although similar definitions can be found on robustness certificates~\cite{DBLP:conf/kdd/ZugnerG19,DBLP:conf/kdd/ZugnerG20,DBLP:conf/nips/BojchevskiG19} on GCNs, our goal is to find a global solution for the entire dataset while the target of previous work is a single node.

\section{Methods}
In this section, we elaborate on how to generate features of the trigger node. Following the previous work~\cite{DBLP:conf/kdd/ZugnerAG18,DBLP:conf/kdd/ZugnerG19}, we consider the situation that the input node features are discrete/binary. Linear Graph Convolution Backdoor is firstly introduced. Then we illustrate how to extend existing graph attacks in our attack scenario, and finally, the time complexity of these methods is presented.
% In this paper, we assume attackers can not change the parameters of the original model. Since attackers activate the backdoor by inserting the trigger node (i.e, connect the trigger node and target node), the major problem is how to generate features of the trigger node.  Two types of backdoor attacks, gradient-based backdoor, and linear graph convolution backdoor, are introduced in the following sections. Figure~\ref{fig:framework} shows the overall framework of our methods.

\subsection{Linear Graph Convolution Backdoor}
The linear approximation method of the problem and its solution are presented as follows.

\begin{figure}[t]
\centering
\includegraphics[width=0.35\textwidth]{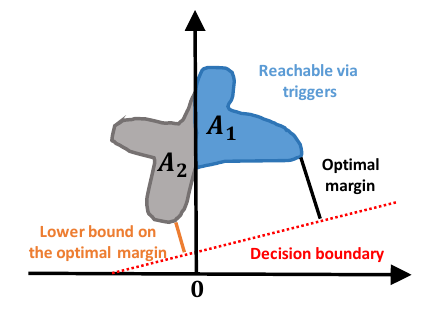}
\caption{A high-level illustration of our linear approximation.} \label{fig:linear}
\end{figure}

\subsubsection{Linear approximation} 
To find an approximated solution of Problem~\ref{pro1}, we remove the ReLU activation function. Figure~\ref{fig:linear} illustrates our core idea. The area $A_1$ denotes all possible points that can be achieved by inserting triggers. As Fig.~\ref{fig:linear} shows, the activation function will discard the area $A_2$ since their horizontal coordinate values are negative. By removing the ReLU function, we search for an approximated solution in a larger space (i.e., areas $A_1$ and $A_2$), which leading to:

\begin{equation}\label{eq:10}
\tilde{\bm{H}}^{(L)} = (\bm{S}^{\prime})^{L}\bm{X}^{\prime}\prod_{l=0}^{L-1}\bm{W}^{(l)} = \tilde{\bm{S}}\bm{X}^{\prime}\tilde{\bm{W}} \ ,
\end{equation}
where $\tilde{\bm{H}}^{(L)}$ is the linear approximation of $\bm{H}^{(L)}$, $\tilde{\bm{S}}$ denotes the $L$-th power of $\bm{S}^{\prime}$ therefore its $v$-th row vector $\tilde{\bm{s}}^{v}$ can represent the $L$-hop neighbors of node $v$, and the weights products are collapsed to a single matrix $\tilde{\bm{W}}\in\mathbb{R}^{d\times k}$ where its $c$ column vector $\tilde{\bm{w}}_{c}\in\mathbb{R}^{d}$ select features that correlated with a specific label class $c$ from all $d$ features. Thus, Eq.~\ref{eq:pro} can be reformulated as follows:
\begin{equation}\label{eq:11}
m_{t, c}^{v} = \min_{\bm{x}^u\in \mathcal{P}_q(\bm{x}^u)}\tilde{\bm{s}}^{v} \bm{X}^{\prime} (\tilde{\bm{w}}_c - \tilde{\bm{w}}_t) \ ,
\end{equation}

Note that model weights (including $\tilde{\bm{w}}_c$ and $\tilde{\bm{w}}_t$) are static, the trigger feature $\bm{x}^u$ can be modified, and the target node's features $\bm{x}^{v}$ can not be changed. Our target is finding an optimal $\bm{x}^u$ that can change the prediction of target node $v$ to a specified label $t$. This is possible since the trigger is the direct neighbor of $v$ and GCNs employ neighbor information to make prediction. Thus, the key point is to determine the trigger's features:

\begin{equation}\label{eq:12}
\begin{aligned}
\bm{x}^u = \mathop{\arg\min}_{\bm{x}^u\in \mathcal{P}_q(\bm{x}^u)}\sum_{c=1}^{k}\tilde{\bm{s}}^{v} \bm{X}^{\prime} (\tilde{\bm{w}}_c - \tilde{\bm{w}}_t) \ , \\
= \mathop{\arg\min}_{\bm{x}^u\in \mathcal{P}_q(\bm{x}^u)}\tilde{\bm{s}}^{v} \bm{X}^{\prime} \sum_{c=1}^{k}(\tilde{\bm{w}}_c - \tilde{\bm{w}}_t) \ ,
\end{aligned}
\end{equation}

Let $\bm{\delta}_t = \sum_{c=1}^{k}(\tilde{\bm{w}}_c - \tilde{\bm{w}}_t) \in \mathbb{R}^{d}$ denote the feature preference difference between target label $t$ and all other labels. Eq.~\ref{eq:12} is equivalent to:
\begin{equation}\label{eq:13}
\min_{\bm{x}^u} \tilde{s}^{v}_{u} \bm{x}^u \bm{\delta}_{t}, \text{ s.t. } \sum_{i=1}^{d} x^u_i = q \text{ and } x^u_i=0 \text{ or } 1 \ ,
\end{equation}
where $\tilde{s}^{v}_{u}$ is positive. Since trigger elements are binary (i.e., 0/1) and the budgets is $q$, such a problem is a zero-one integer linear programming. A greedy solution can be applied to find the optimal solution in linear complexity.

% TODO: Proposition - Proof

\subsubsection{Solution} Suppose $\mathcal{C}=\{d_1, d_2, \dots, d_q\}$ a combination which is a subset of $\{1, 2, \dots, d\}$ of size $q$, where $1 \leq q \leq d$ and $1 \leq d_1, d_2, \dots, d_q \leq d$, an assignment of $\bm{x}^u$ by $\mathcal{C}$ is defined as:

\begin{equation}\label{eq:14}
\begin{aligned}
x^u_i = 
    \left\{
    \begin{array}{cc}
        1 \ , & \ i \in \mathcal{C} \ , \\
        0 \ , & \ i \notin \mathcal{C} \ ,
    \end{array}
    \right.
\end{aligned}
\end{equation}
where $\bm{x}^u$ satisfy all constrains, substitute back to Eq.~\ref{eq:13} we have:

\begin{equation}\label{eq:15}
\min_{\bm{x}^u} \tilde{s}^{v}_{u} \sum_{i=1}^{q} \delta_{t}^{d_i} \ ,
\end{equation}

\begin{figure}[t]
\centering
\includegraphics[width=0.45\textwidth]{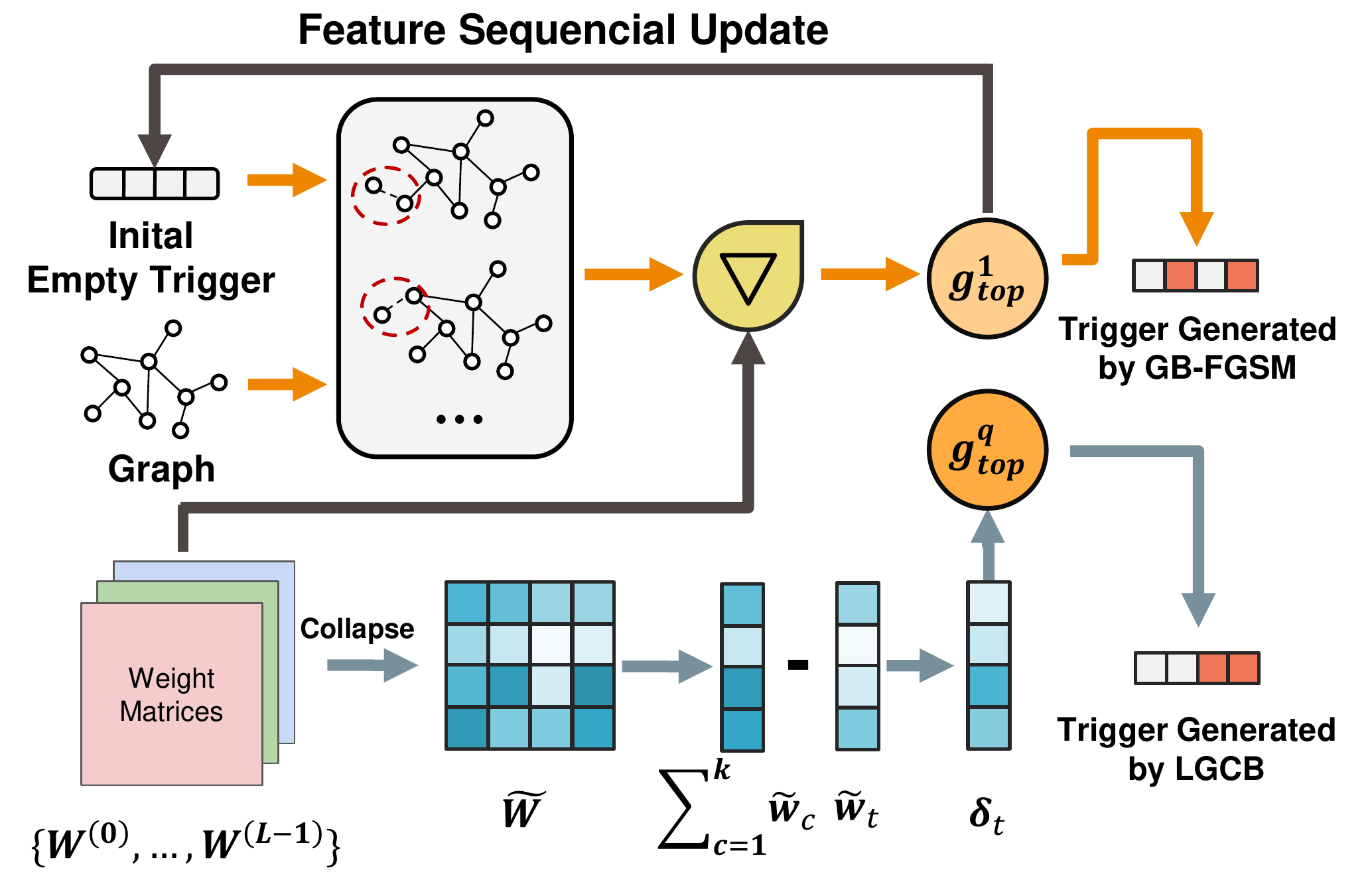}
\caption{The framework of backdoor attacks on node classification. The top row is the procedure of GB-FGSM as an example of Gradient-based Backdoor. Firstly, the initial empty trigger is activated in the input graph by connecting with the target node that is specified by the attacker. The weight matrices at the left bottom corner represent model parameters, $\nabla$ operator is used to compute gradients w.r.t. trigger, and the $g_{\text{top}}^{1}$ operator means generating top-1 feature from gradients, the generated feature then updates in the trigger and starts next iteration. After $q$ iterations, the trigger is generated by GB-FGSM. The bottom row is LGCB that only utilize model parameters, the shade of color blocks corresponds to the magnitude of parameters. $g_{\text{top}}^{q}$ operator select top-$q$ features to generate LGCB trigger.} \label{fig:framework}
\end{figure}

% \begin{algorithm}[t]
%     \KwIn{model weights $\mathcal{\bm{W}}$, budget $q$, target label $t$}
%     \KwOut{trigger $\bm{x}^u$ }
%     $\tilde{\bm{W}} \leftarrow \bm{W}^{(0)}\cdots\bm{W}^{(L-1)}$\;
%     $\bm{\delta}_t \leftarrow \sum_{c=1}^{k}(\tilde{\bm{w}}_c - \tilde{\bm{w}}_t)$\;
%     $\bm{x}^{u} \leftarrow g_{\text{top}}(-\bm{\delta}_t;q)$\;
%     return $\bm{x}^u$\;
%     \caption{{\bf Linear Graph Convolution Backdoor} \label{alg:lgcb}}
% \end{algorithm}

%  If any index $d_i$ is in top $q$ minimal elements but not in $\{d_1, d_2, \dots, d_q\}$, we can replace $\mathop{\arg\max}_{d_i} \delta_t^{d_i}$ with $d_i$. Finally, we use the top $q$ minimal element indexes of $\bm{\delta}_{t}$ to assign the optimal trigger by Eq.~\ref{eq:14}. 

Eq.~\ref{eq:15} get minimal point if $\mathcal{C}$ is the indexes of top $q$ minimal elements of $\bm{\delta}_{t}$ (or top $q$ maximal elements of $-\bm{\delta}_{t}$). So the solution of Pro.~\ref{pro1} is derived as:

\begin{equation}\label{eq:16}
\bm{x}^{u} = g_{\text{top}}(-\bm{\delta}_{t};q) \ ,
\end{equation}

Note that our solution is determined only by $\bm{\delta}_{t}$ and not related with $\tilde{s}^{v}_{u}$. Therefore, LGCB can generate trigger using less knowledge, while still achieve competitive attack performance (show in Table~\ref{table:success}) compared to methods using full knowledge. In other words, no matter which node $v$ is targeted to attack, we have the same closed-form solution given a specific target label $t$. Compare to other targeted attack methods that usually construct a particular trigger for each different node, our trigger can perform a generally backdoor attack towards all nodes in the graph. 

% Our trigger is generated with the only limited knowledge $\mathcal{W}$, while the intrinsic relation between features and labels from dataset $\mathcal{D}$ which has been embedded in model weights $\mathcal{W}$ can be inferred by our method without knowing $\mathcal{D}$, such that it's still competitive when the linear optimal solution transferred to the non-linear situation.

\subsection{Extending Existing Graph Attacks}
Existing graph attack methods usually utilize gradients to find a local solution (i.e., a unique solution for each node) while our backdoor attacks require a global solution (i.e., trigger) for all nodes. In this section, we show that how to extend these models to obtain a global trigger.

\subsubsection{Overall Framework}

From an optimization perspective, existing models find an optimal trigger to minimize the attack loss in a given perturbation space. Their key contribution is the approach to utilize gradient information to flip components (in discrete data domain) of graph structure or node features. Thus, by adopting the same methods, it is feasible to optimize the binary trigger features. Following the general routine of existing backdoor attacks, a contaminated dataset is firstly prepared by connecting the triggers (whose features are all zero) to each training nodes. Then FGSM, PGD, and IG are employed to generate the trigger features. Figure~\ref{fig:framework} displays an example of our framework.

\paragraph{Notation.} We briefly introduce the notations used in this section. For a specific target node $v$, let $h^{v}(\bm{x}^{u})\in\mathbb{R}^{k}$ denote the logits output of node $v$ with a connected trigger $\bm{x}^{u}$, $l:\mathbb{R}^k\to\mathbb{R}$ denote the attack loss which is the negative cross-entropy loss w.r.t. target label $t$ for node $v$:

\begin{equation}
    l(h^{v}(\bm{x}^{u});t)=-\ln[h^{v}(\bm{x}^{u})]_{t}-\sum_{c\neq t}\ln[h^{v}(\bm{x}^{u})]_{c}
\end{equation}
Note that the attack loss depends on the trigger thus its gradients can be derived by chaining rules. Let $\gamma_i=\partial{l}/\partial{x^{u}_{i}}$ denotes the gradient of trigger's $i$-th entry. Given a gradient vector $\bm{\gamma}\in\mathbb{R}^{d}$ and budget $q$, a generator function $g:\mathbb{R}^{d}\to\{0,1\}^d$ is needed to generate trigger $\bm{x}^{u}=g(\bm{\gamma};q)$ from gradients. Since the gradient can reflect the importance of corresponding feature, we define a top generator $g_{\text{top}}(\cdot)$ that assign top $q$ features with largest gradients. % TODO: define g_top

\subsubsection{Algorithm}
In the following, three gradient-based attack methods (including GB-FGSM, GB-PGD, and GB-IG) are presented to generate the trigger features. 

\paragraph{GB-FGSM} 

\begin{algorithm}[t]
    \KwIn{model weights $\mathcal{\bm{W}}$, graph $\mathcal{G}=(\bm{A}, \bm{X})$, target node $v$, target label $t$, budget $q$}
    \KwOut{trigger features $\bm{x}^{u}$}
    $\bm{x}^{u} \leftarrow \bm{0}$\;
    % $\tilde{\bm{W}} \leftarrow \bm{W}^{(0)}\cdots\bm{W}^{(k-1)}$\;
    \For{$p$ = $1$ to $q$} {
    
        $\bm{\gamma}^{(p)} \leftarrow \frac{\partial{l(h^{v}(\bm{x}^{u});t)}}{\partial{\bm{x}^{u}}} $\;
        $\bm{\gamma}^{(p)} \leftarrow \bm{\gamma}^{(p)}*(1-\bm{x}^{u})$; \ // mask flipped features \\
        $\bm{x}^{u} \leftarrow \bm{x}^{u} + g_{\text{top}}(\bm{\gamma}^{(p)};1)$\;
    }
    return $\bm{x}^{u}$\;
    \caption{{\bf Gradient-based Backdoor: FGSM} \label{alg:fgsm}}
\end{algorithm}

In computer vision, Fast Gradient Sign Method (FGSM) \cite{DBLP:journals/corr/GoodfellowSS14} generates adversarial examples by adding imperceptible change along the gradient direction of each pixel in images. For targeted attack, FGSM minimize the target loss function in a one-step gradient update:

\begin{equation}\label{eq:4}
\bm{x}^{u}=[\bm{x}^{u}]^{(0)} - \epsilon\text{sign}(\frac{\partial{l(h^{v}([\bm{x}^{u}]^{(0)});t)}}{\partial{\bm{x}^{u}}}) \ , \ [\bm{x}^{u}]^{(0)}=\bm{0}
\end{equation}
Although such a one-step FGSM is simple and fast for continuous adversarial examples, it is not feasible for trigger generation in the discrete graph domain. The reason is that the generated adversarial examples are satisfies the max-norm constraint $||\bm{x}^{u}-[\bm{x}^{u}]^{(0)}||_{\infty}\leqslant\epsilon$, which lies in $\epsilon$-neighbor ball of the initial empty trigger $[\bm{x}^{u}]^{(0)}$. If $\epsilon$ is too small, the $\epsilon$-ball won't contain any discrete point. On the contrary, if it's too large, the gradient effect would be unpredictable because of the model's non-linearity. To cope with this dilemma, an extended version GB-FGSM, which uses a greedy step-by-step rather than one-step optimization, is proposed to expand FGSM in the graph domain. Algorithm \ref{alg:fgsm} shows the pseudo-code of GB-FGSM for discrete trigger generation. Each step we only select one most promising feature among all features by $g_{\text{top}}(\bm{\gamma}^{(p)};1)$ and add it to the output trigger. Then the updated trigger is passed to the next gradient iteration. To prevent flipping one feature multiple times, the gradients are masked by the trigger before selection. The trigger is iteratively updated until the budget $q$ is exhausted.

\paragraph{GB-PGD} 

\begin{algorithm}[t]
    \KwIn{model weights $\mathcal{\bm{W}}$, graph $\mathcal{G}=(\bm{A}, \bm{X})$, target node $v$, target label $t$, budget $q$, learning rate $\eta_p$, PGD steps $T$, sampling steps $K$}
    \KwOut{trigger features $\bm{x}^{u}$}
    $[\bm{x}^{u}]^{(0)} \leftarrow \bm{0}$\;
    % $\tilde{\bm{W}} \leftarrow \bm{W}^{(0)}\cdots\bm{W}^{(k-1)}$\;
    \For{$p$ = $1$ to $T$} {
        $\bm{\gamma}^{(p)} \leftarrow \frac{\partial{l(h^{v}([\bm{x}^{u}]^{(p-1)});t)}}{\partial{\bm{x}^{u}}} $\;
        $[\bm{x}^{u}]^{(p)} \leftarrow [\bm{x}^{u}]^{(p-1)} - \eta_{p}\bm{\gamma}^{(p)}$\;
        call projection operation for $[\bm{x}^{u}]^{(p)}$ by Equation~(\ref{eq:6})
    }
    $\bm{x}^{u} \leftarrow \bm{0}$\;
    \For{$k$ = $1$ to $K$} {
        call random generator $[\bm{x}^{u}]^{[k]} \leftarrow g_{\text{rnd}}([\bm{x}^{u}]^{(T)})$ \;
        \If{$\bm{1}^{T}[\bm{x}^{u}]^{[k]} \leqslant q$} {
            \If{$l(h^{v}([\bm{x}^{u}]^{[k]});t) < l(h^{v}(\bm{x}^{u});t)$} {
                $\bm{x}^{u} \leftarrow [\bm{x}^{u}]^{[k]}$\;
            }
        }
    }
    return $\bm{x}^{u}$\;
    \caption{{\bf Gradient-based Backdoor: PGD} \label{alg:pgd}}
\end{algorithm}

As we see in FGSM, although there are discrete points feasible and an optimal point exists when $\epsilon$ is large enough, it's hard to find it in a one-step optimization. Projected Gradient Descent (PGD) \cite{DBLP:conf/iclr/MadryMSTV18} is a multi-step variant of FGSM and has been exploited for topology attack in the graph domain. In this work, it is extended to feature attack (which is binary as well) to generate triggers. First, to directly employ gradients on each step, the trigger $\bm{x}^{u}\in\{0,1\}^{d}$ is relaxed to $\bm{x}^{u}\in[0,1]^{d}$, thus the perturbation space relaxed to a continuous space $\mathcal{P}=\{\bm{x}^{u} | \bm{1}^T \bm{x}^{u} \leq\epsilon, \bm{x}^{u}\in[0,1]^{d}\}$. We then optimize attack loss w.r.t. relaxed $\bm{x}^{u}$ by PGD:

\begin{equation}\label{eq:5}
[\bm{x}^{u}]^{(p)}=\Pi_{\mathcal{P}}\left( [\bm{x}^{u}]^{(p-1)} - \eta_{p}\bm{\gamma}^{(p)} \right) \ ,
% \bm{\gamma}^{(p)}=\frac{\partial{l(h^{v}([\bm{x}^{u}]^{(p-1)});t)}}{\partial{\bm{x}^{u}}} \ ,
\end{equation}
where $[\bm{x}^{u}]^{(p)}$ denotes the trigger in step $p$, which is iteratively updated by gradients $\bm{\gamma}^{(p)}$. $\eta_p$ is the learning rate, and $\Pi_{\mathcal{P}}(\bm{x}^{u})=\arg\min_{\bm{x}\in\mathcal{P}}\lVert\bm{x}-\bm{x}^{u}\rVert_{2}^{2}$ is the projection operator to ensure $\bm{x}^{u}\in\mathcal{P}$ during optimization process, which has a closed-form solution:

\begin{equation}\label{eq:6}
\begin{aligned}
\Pi_{\mathcal{P}}(\bm{x}) = 
    \left\{
    \begin{array}{ll}
        P_{[0,1]}[\bm{x}-\mu\bm{1}] \ ,  & \text{ if }
            \mu>0 \text{ and } \bm{1}^T P_{[0,1]}[\bm{x}-\mu\bm{1}]=\epsilon \ , \\
        P_{[0,1]}[\bm{x}] \ , & \text{ if }
            \bm{1}^T P_{[0,1]}[\bm{x}]\leqslant\epsilon\ ,
    \end{array}
    \right.
\end{aligned}
\end{equation}
where $P_{[0,1]}(x)=\max(0, \min(1, x))$ is the clipping operator.
This projection operation moves the out-of-constraint trigger back to the perturbation space while the relative order between features preserved. The proof can be found in \cite{DBLP:conf/ijcai/XuC0CWHL19}. After $T$ PGD iterations, we obtain a continuous trigger $[\bm{x}^{u}]^{(T)}\in[0,1]^{d}$ and now need to convert it to a binary trigger $\bm{x}^{u}\in\{0,1\}^{d}$. If we treat $[\bm{x}^{u}]^{(T)}$ as a stochastic feature vector, a new generator function $g_{\text{rnd}}(\cdot)$ can be defined based on random sampling from these stochastic features:

\begin{equation}\label{eq:7}
g_{\text{rnd}}\left([\bm{x}^{u}]^{(T)}\right)=\mathbb{I}\left([\bm{x}^{u}]^{(T)}>\bm{r}\right) \ ,
\end{equation}
where $\mathbb{I}$ is the indicator function, $\bm{r}$ is a uniform random vector that $r_i\sim^{\text{i.i.d.}}U[0,1]$. In this way, the features with larger gradients are more likely occurring in the output trigger. Consider uncertainty of the process, we set $\epsilon=q$, then evaluate it $K$ times and select one trigger $[\bm{x}^{u}]^{[k]}$ which yields the lowest loss under $q$ budget. The overall procedure is summarized in Algorithm~\ref{alg:pgd}.

\paragraph{GB-IG} 
FGSM shows that using local gradients to optimize discrete variables is inaccurate. To address this problem, an integrated gradients based method (GB-IG)~\cite{DBLP:conf/ijcai/Wu0TDLZ19} is proposed. Under the discrete setting, the integrated gradient is defined as follows: given a objective $l:\mathbb{R}^{k}\to\mathbb{R}$ target to label $t$, let trigger $\bm{x}^{u}\in\mathbb{R}^{d}$ be the input, all-one trigger $\bm{1}^{d}$ is the baseline input. Consider a straight line path between $\bm{x}^{u}$ and $\bm{1}^{d}$, the integrated gradients of $l$ w.r.t. $\bm{x}^{u}$ is obtained by uniformly sampling $m$ points on this path and accumulating gradients of each point:

\begin{equation}\label{eq:8}
\gamma_{i} \approx (1-x^{u}_{i}) \times \sum_{j=1}^{m} \frac{ \partial{l(h^{v}(\frac{j}{m} \times (\bm{1}-\bm{x}^{u}));t)} }  { \partial{x^{u}_{i}} } \times \frac{1}{m} \ ,
\end{equation}

\begin{algorithm}[t]
    \KwIn{model weights $\mathcal{\bm{W}}$, graph $\mathcal{G}=(\bm{A}, \bm{X})$, target node $v$, target label $t$, budget $q$, steps $m$}
    \KwOut{trigger features $\bm{x}^{u}$}
    $\bm{x}^{u} \leftarrow \bm{0}$\;
    % $\tilde{\bm{W}} \leftarrow \bm{W}^{(0)}\cdots\bm{W}^{(k-1)}$\;
    \For{$i$ = $1$ to $d$} {
        calculate $\gamma_{i}$ by Equation (\ref{eq:8})\;
    }
    // generate trigger with $q$ largest integrated gradients \\
    $\bm{x}^{u} \leftarrow g_{\text{top}}(\bm{\gamma};q)$\;
    return $\bm{x}^{u}$\;
    \caption{{\bf Gradient-based Backdoor: IG} \label{alg:ig}}
\end{algorithm}

Algorithm~\ref{alg:ig} shows the procedure of IG applied to triggers. In our scenario, with an initial empty trigger, only feature addition is needed. We compute and accumulate the gradients of target loss $l$ w.r.t. trigger entries $\bm{x}^{u}$. The integrated gradients are considered as the importance of each trigger feature, and $q$ largest gradients are selected to assign the trigger who has the most probability to minimize target loss within a given budget.

\subsection{Unnoticeable generator function}

The generator functions used in LGCB and the gradient-based backdoors are simply retrieving the most important features from the preference difference vector or the gradient vectors, while not considering the stealthiness of these features which is one of the attackers' goals. So a unnoticeable generator function is needed to ensure the noticeable features w.r.t. the target node would not be selected in the trigger node. Follow the feature unnoticeable perturbation method in \cite{DBLP:conf/kdd/ZugnerAG18}, it is noticeable if two features suddenly occurred together which never co-occurring in the original graph. Let $\bm{E}\in\{0,1\}^{d\times d}$ be the feature co-occurrence matrix of the original graph, we force that all the features in the trigger node must have been occurred with one of the features of the target node at least once. The features that never co-occurring with the features of the target node would be filtered by the unnoticeable generator function:

\begin{equation}
g_{ste}(\bm{\gamma};q) = g_{top}\left(\bm{\gamma} \cdot \mathbb{I}\left(\sum_{i=1}^{d}\bm{E}[x^v_i] > 0\right);q\right) \ ,
\end{equation}
where $\bm{E}^v$ denote the co-occurred features of the target node $v$, $\mathbb{I}$ is the indicator function, and the top generator can be replaced as other generators as well.

% TODO: Summarize Gradient-based Backdoor
\subsection{Time complexity}
To compare the time complexity of the aforementioned approaches, a two-layer GCN is considered as the target model, where $d$, $h$, $k$ is the count of input units, hidden units, and output units, respectively. For the gradient-based backdoors, the gradient computation including feed-forward and backward propagation, thus its time complexity is $O(edhk)$, where $e$ is the edge count in the graph~\cite{DBLP:conf/iclr/KipfW17}.
The complexities of gradient-based backdoors are mainly contributed by the gradient iteration steps. The complexity of GB-FGSM is $O(qedhk)$ where $q$ is the budget, GB-PGD is $O((T+K)edhk)$ where $T$ and $K$ are the training and random sampling steps, GB-IG is $O(medhk)$ where $m$ is the sampling steps.
% The time complexities of the proposed backdoors are listed in Table~\ref{table:
Note that the time complexity of GB-PGD and GB-IG is irrelevant to budget $q$, and only depend on their hyper-parameter settings and the scale of the graph. LGCB's time complexity is $O(dhk+q\log{d})$ which is mainly contributed by the weight matrix collapse and the top-$q$ selector operation. The matrix collapse operation only requires complexity $O(dhk)$, where $h$ and $k$ are usually small constant factors, therefore scales linearly in the number of the features. 
% Further, the matrix collapse operation can be processed in a pre-computed way with the given model parameters, such that the time complexity of LGCB is improved to $O(q\log{d})$.

% \begin{table}
% \small
% \caption{The time complexities of the proposed backdoors.} \label{table:complexity}
% \setlength{\tabcolsep}{2.0mm}
% \centering
% % \begin{tabular}{@{}c|p{2cm}<{\centering}|p{2cm}<{\centering}|p{2cm}<{\centering}|p{2cm}<{\centering}@{}}
% \begin{tabular}{@{}p{1.6cm}<{\centering}|c@{}}
% \toprule
% \textbf{Approach} & \textbf{Time Complexity} \\ 
% \midrule 
% \textbf{GB-FGSM} & $O(qedhk)$ \\
% \textbf{GB-PGD} & $O((T+K)edhk)$ \\
% \textbf{GB-IG} & $O(medhk)$ \\
% \textbf{LGCB} & $O(dhk+q\log{d})$ \\
% \bottomrule
% \end{tabular}
% \end{table}

\section{Experiments}
In this section, we conduct experiments on four real-world datasets aiming to answer the following research questions:

\begin{description}
% 	\item[RQ1] Given trained GCNs, can the proposed LGCB and gradient-based methods effectively perform backdoor attacks on node classification tasks? 
% 	\item[RQ2] How are the running times of different backdoors (i.e., LGCB and gradient-based methods)?
% 	\item[RQ3] How does the performance of backdoors vary as the degree of target nodes?
	\item[RQ1] Given trained GCNs, can the proposed LGCB and gradient-based methods effectively perform backdoor attacks on node classification tasks? 
	\item[RQ2] Can the designed trigger node be easily detected and further circumvented?
	\item[RQ3] How the trigger connectivities affect the attack performance of the proposed backdoors?
\end{description}

% \begin{figure*}[t]
%   \centering
%   \includegraphics[width=0.24\textwidth]{blog-100k.pdf}
%   \includegraphics[width=0.24\textwidth]{blog-200k.pdf}
%   \includegraphics[width=0.24\textwidth]{blog-300k.pdf}
%   \includegraphics[width=0.24\textwidth]{blog-400k.pdf}
%   \includegraphics[width=0.24\textwidth]{flickr-10k.pdf}
%   \includegraphics[width=0.24\textwidth]{flickr-50k.pdf}
%   \includegraphics[width=0.24\textwidth]{flickr-100k.pdf}
%   \includegraphics[width=0.24\textwidth]{flickr-200k.pdf}
%   \includegraphics[width=0.24\textwidth]{cora-10k.pdf}
%   \includegraphics[width=0.24\textwidth]{cora-50k.pdf}
%   \includegraphics[width=0.24\textwidth]{cora-100k.pdf}
%   \includegraphics[width=0.24\textwidth]{cora-200k.pdf}
%   \includegraphics[width=0.24\textwidth]{pubmed-1k.pdf}
%   \includegraphics[width=0.24\textwidth]{pubmed-10k.pdf}
%   \includegraphics[width=0.24\textwidth]{pubmed-30k.pdf}
%   \includegraphics[width=0.24\textwidth]{pubmed-100k.pdf}
%   \caption{Overall Comparison.} \label{fig:group}
% \end{figure*}

\begin{table}
\small
\caption{The statistics of datasets and the accuracy of 2-layer GCNs. The results are averaged over 5 runs.} \label{table:data}
\setlength{\tabcolsep}{2mm}
\centering
% \begin{tabular}{@{}c|p{2cm}<{\centering}|p{2cm}<{\centering}|p{2cm}<{\centering}|p{2cm}<{\centering}@{}}
\resizebox{0.48\textwidth}{!}{
\begin{tabular}{@{}p{2.0cm}<{\centering}|c|c|c|c@{}}
\toprule
\textbf{Dataset} & \textbf{BlogCatalog} & \textbf{Flickr} & \textbf{Cora} & \textbf{PubMed} \\ 
\midrule 
\textbf{\#Node} & 5,196 & 7,575 &  2,485 & 19,717\\
\textbf{\#Edge} & 171,743 & 239,738 &  5,069 & 44,324 \\
\textbf{\#Feature} & 8,189 & 12,047 & 1,433 & 500\\
\textbf{\#Class} & 6 & 9 &  7 & 3\\
\textbf{Accuracy (\%)} & 72.42$\pm$0.54 & 60.14$\pm$0.22 & 84.27$\pm$0.32 & 85.00 $\pm$0.21 \\
\bottomrule
\end{tabular}
}
\end{table}

\begin{table*}
\caption{The attack success rate comparison of backdoors. $t$ represents the target label and the results are averaged over 5 runs. GB-FGSM performs best. The performance of GB-PGD and LGCB is compatible. IG achieves the worst performance.} \label{table:success}
\small
\setlength{\tabcolsep}{4mm}
\centering
% \begin{tabular}{@{}c|p{2cm}<{\centering}|p{2cm}<{\centering}|p{2cm}<{\centering}|p{2cm}<{\centering}@{}}
\resizebox{\textwidth}{!}{
\begin{tabular}{@{}p{1.5cm}<{\centering}p{0.95cm}<{\centering}p{0.95cm}<{\centering}p{0.95cm}<{\centering}p{0.95cm}<{\centering}p{0.95cm}<{\centering}p{0.95cm}<{\centering}p{0.95cm}<{\centering}p{0.95cm}<{\centering}p{0.95cm}<{\centering}p{0.95cm}<{\centering}p{0.95cm}<{\centering}p{1.2cm}<{\centering}@{}}
\toprule
\multirow{2}{*}{\textbf{BlogCatalog}}  & \multicolumn{3}{c}{\textbf{GB-FGSM}} & \multicolumn{3}{c}{\textbf{GB-IG}} & \multicolumn{3}{c}{\textbf{GB-PGD}} & \multicolumn{3}{c}{\textbf{LGCB}}\\ 
\cmidrule{2-13}
& q=200 & q=300 & q=400  & q=200 & q=300 & q=400 & q=200 & q=300 & q=400 & q=200 & q=300 & q=400\\
\midrule
t=1 & 0.92$\pm$0.03 & 0.96$\pm$0.02 & 0.98$\pm$0.01 & 0.91$\pm$0.04 & 0.95$\pm$0.03 & 0.97$\pm$0.01 & 0.92$\pm$0.03 & 0.96$\pm$0.02 & 0.98$\pm$0.01 & 0.92$\pm$0.03 & 0.96$\pm$0.02 & 0.98$\pm$0.01 \\
t=2 & 0.94$\pm$0.04 & 0.97$\pm$0.02 & 0.98$\pm$0.01 & 0.93$\pm$0.04 & 0.97$\pm$0.03 & 0.98$\pm$0.02 & 0.94$\pm$0.04 & 0.97$\pm$0.02 & 0.98$\pm$0.01 & 0.93$\pm$0.04 & 0.97$\pm$0.02 & 0.98$\pm$0.01 \\
t=3 & 0.86$\pm$0.08 & 0.92$\pm$0.05 & 0.95$\pm$0.04 & 0.85$\pm$0.08 & 0.91$\pm$0.06 & 0.94$\pm$0.05 & 0.86$\pm$0.08 & 0.92$\pm$0.05 & 0.94$\pm$0.04 & 0.85$\pm$0.08 & 0.91$\pm$0.06 & 0.94$\pm$0.05 \\
t=4 & 0.95$\pm$0.03 & 0.98$\pm$0.02 & 0.99$\pm$0.01 & 0.95$\pm$0.03 & 0.98$\pm$0.02 & 0.99$\pm$0.01 & 0.95$\pm$0.03 & 0.98$\pm$0.02 & 0.99$\pm$0.01 & 0.95$\pm$0.03 & 0.98$\pm$0.02 & 0.99$\pm$0.01 \\
t=5 & 0.83$\pm$0.06 & 0.90$\pm$0.04 & 0.94$\pm$0.03 & 0.82$\pm$0.07 & 0.89$\pm$0.05 & 0.93$\pm$0.03 & 0.82$\pm$0.06 & 0.90$\pm$0.04 & 0.93$\pm$0.03 & 0.82$\pm$0.07 & 0.89$\pm$0.05 & 0.93$\pm$0.04 \\
t=6 & 0.92$\pm$0.04 & 0.96$\pm$0.03 & 0.98$\pm$0.01 & 0.91$\pm$0.04 & 0.95$\pm$0.03 & 0.98$\pm$0.02 & 0.92$\pm$0.04 & 0.96$\pm$0.03 & 0.98$\pm$0.02 & 0.91$\pm$0.04 & 0.96$\pm$0.03 & 0.98$\pm$0.02 \\
\midrule
\midrule
\multirow{2}{*}{\textbf{Flickr}}  & \multicolumn{3}{c}{\textbf{GB-FGSM}} & \multicolumn{3}{c}{\textbf{GB-IG}} & \multicolumn{3}{c}{\textbf{GB-PGD}} & \multicolumn{3}{c}{\textbf{LGCB}}\\ 
\cmidrule{2-13}
& q=50 & q=100 & q=200  & q=50 & q=100 & q=200 & q=50 & q=100 & q=200 & q=50 & q=100 & q=200\\
\midrule
t=1 & 0.90$\pm$0.01 & 0.96$\pm$0.01 & 0.98$\pm$0.00 & 0.88$\pm$0.01 & 0.95$\pm$0.01 & 0.98$\pm$0.00 & 0.89$\pm$0.01 & 0.96$\pm$0.01 & 0.98$\pm$0.00 & 0.89$\pm$0.01 & 0.95$\pm$0.01 & 0.98$\pm$0.00 \\
t=2 & 0.95$\pm$0.01 & 0.98$\pm$0.00 & 0.99$\pm$0.00 & 0.94$\pm$0.01 & 0.98$\pm$0.01 & 0.99$\pm$0.00 & 0.95$\pm$0.01 & 0.98$\pm$0.00 & 0.99$\pm$0.00 & 0.95$\pm$0.01 & 0.98$\pm$0.00 & 0.99$\pm$0.00 \\
t=3 & 0.95$\pm$0.01 & 0.98$\pm$0.01 & 1.00$\pm$0.00 & 0.95$\pm$0.02 & 0.98$\pm$0.01 & 0.99$\pm$0.00 & 0.95$\pm$0.01 & 0.98$\pm$0.01 & 0.99$\pm$0.00 & 0.95$\pm$0.02 & 0.98$\pm$0.01 & 1.00$\pm$0.00 \\
t=4 & 0.96$\pm$0.01 & 0.99$\pm$0.00 & 1.00$\pm$0.00 & 0.95$\pm$0.01 & 0.98$\pm$0.00 & 1.00$\pm$0.00 & 0.96$\pm$0.01 & 0.98$\pm$0.00 & 1.00$\pm$0.00 & 0.95$\pm$0.01 & 0.98$\pm$0.00 & 1.00$\pm$0.00 \\
t=5 & 0.97$\pm$0.01 & 0.99$\pm$0.00 & 1.00$\pm$0.00 & 0.97$\pm$0.01 & 0.99$\pm$0.00 & 1.00$\pm$0.00 & 0.97$\pm$0.01 & 0.99$\pm$0.00 & 1.00$\pm$0.00 & 0.97$\pm$0.01 & 0.99$\pm$0.00 & 1.00$\pm$0.00 \\
t=6 & 0.97$\pm$0.00 & 0.99$\pm$0.00 & 1.00$\pm$0.00 & 0.96$\pm$0.01 & 0.99$\pm$0.00 & 1.00$\pm$0.00 & 0.97$\pm$0.00 & 0.99$\pm$0.00 & 1.00$\pm$0.00 & 0.96$\pm$0.01 & 0.99$\pm$0.00 & 1.00$\pm$0.00 \\
t=7 & 0.94$\pm$0.00 & 0.98$\pm$0.00 & 0.99$\pm$0.00 & 0.93$\pm$0.00 & 0.97$\pm$0.00 & 0.99$\pm$0.00 & 0.94$\pm$0.00 & 0.98$\pm$0.00 & 0.99$\pm$0.00 & 0.94$\pm$0.00 & 0.98$\pm$0.00 & 0.99$\pm$0.00 \\
t=8 & 0.94$\pm$0.01 & 0.98$\pm$0.01 & 0.99$\pm$0.00 & 0.94$\pm$0.01 & 0.98$\pm$0.01 & 0.99$\pm$0.00 & 0.94$\pm$0.01 & 0.98$\pm$0.01 & 0.99$\pm$0.00 & 0.94$\pm$0.01 & 0.98$\pm$0.01 & 0.99$\pm$0.00 \\
t=9 & 0.94$\pm$0.01 & 0.98$\pm$0.00 & 0.99$\pm$0.00 & 0.93$\pm$0.01 & 0.97$\pm$0.01 & 0.99$\pm$0.00 & 0.94$\pm$0.01 & 0.98$\pm$0.00 & 0.99$\pm$0.00 & 0.93$\pm$0.01 & 0.98$\pm$0.00 & 0.99$\pm$0.00 \\
\midrule
\midrule
\multirow{2}{*}{\textbf{Cora}}  & \multicolumn{3}{c}{\textbf{GB-FGSM}} & \multicolumn{3}{c}{\textbf{GB-IG}} & \multicolumn{3}{c}{\textbf{GB-PGD}} & \multicolumn{3}{c}{\textbf{LGCB}}\\ 
\cmidrule{2-13}
& q=50 & q=100 & q=200  & q=50 & q=100 & q=200 & q=50 & q=100 & q=200 & q=50 & q=100 & q=200\\
\midrule
t=1 & 0.86$\pm$0.01 & 0.96$\pm$0.00 & 0.99$\pm$0.00 & 0.84$\pm$0.01 & 0.95$\pm$0.00 & 0.98$\pm$0.00 & 0.86$\pm$0.01 & 0.96$\pm$0.00 & 0.99$\pm$0.00 & 0.85$\pm$0.01 & 0.96$\pm$0.01 & 0.99$\pm$0.00 \\
t=2 & 0.91$\pm$0.04 & 0.98$\pm$0.01 & 0.99$\pm$0.00 & 0.89$\pm$0.05 & 0.97$\pm$0.02 & 0.99$\pm$0.00 & 0.91$\pm$0.04 & 0.98$\pm$0.01 & 0.99$\pm$0.00 & 0.90$\pm$0.04 & 0.98$\pm$0.01 & 0.99$\pm$0.00 \\
t=3 & 0.97$\pm$0.01 & 0.99$\pm$0.00 & 1.00$\pm$0.00 & 0.97$\pm$0.01 & 0.99$\pm$0.00 & 1.00$\pm$0.00 & 0.97$\pm$0.01 & 0.99$\pm$0.00 & 1.00$\pm$0.00 & 0.97$\pm$0.01 & 0.99$\pm$0.00 & 1.00$\pm$0.00 \\
t=4 & 0.94$\pm$0.01 & 0.99$\pm$0.00 & 1.00$\pm$0.00 & 0.93$\pm$0.02 & 0.98$\pm$0.00 & 0.99$\pm$0.00 & 0.94$\pm$0.01 & 0.99$\pm$0.00 & 1.00$\pm$0.00 & 0.93$\pm$0.02 & 0.99$\pm$0.00 & 1.00$\pm$0.00 \\
t=5 & 0.84$\pm$0.04 & 0.96$\pm$0.01 & 0.99$\pm$0.00 & 0.80$\pm$0.04 & 0.94$\pm$0.02 & 0.97$\pm$0.02 & 0.84$\pm$0.04 & 0.96$\pm$0.01 & 0.99$\pm$0.00 & 0.82$\pm$0.04 & 0.95$\pm$0.01 & 0.98$\pm$0.00 \\
t=6 & 0.68$\pm$0.06 & 0.89$\pm$0.04 & 0.96$\pm$0.02 & 0.61$\pm$0.09 & 0.80$\pm$0.10 & 0.78$\pm$0.19 & 0.68$\pm$0.06 & 0.89$\pm$0.04 & 0.96$\pm$0.01 & 0.65$\pm$0.06 & 0.87$\pm$0.05 & 0.94$\pm$0.03 \\
t=7 & 0.90$\pm$0.02 & 0.98$\pm$0.01 & 0.99$\pm$0.00 & 0.89$\pm$0.03 & 0.97$\pm$0.01 & 0.99$\pm$0.00 & 0.90$\pm$0.02 & 0.98$\pm$0.01 & 0.99$\pm$0.00 & 0.90$\pm$0.03 & 0.97$\pm$0.01 & 0.99$\pm$0.00 \\
\midrule
\midrule
\multirow{2}{*}{\textbf{PubMed}}  & \multicolumn{3}{c}{\textbf{GB-FGSM}} & \multicolumn{3}{c}{\textbf{GB-IG}} & \multicolumn{3}{c}{\textbf{GB-PGD}} & \multicolumn{3}{c}{\textbf{LGCB}}\\ 
\cmidrule{2-13}
& q=1 & q=10 & q=30  & q=1 & q=10 & q=30 & q=1 & q=10 & q=30 & q=1 & q=10 & q=30\\
\midrule
t=1 & 0.71$\pm$0.04 & 0.98$\pm$0.01 & 1.00$\pm$0.00 & 0.62$\pm$0.05 & 0.98$\pm$0.01 & 1.00$\pm$0.00 & 0.70$\pm$0.05 & 0.98$\pm$0.01 & 1.00$\pm$0.00 & 0.73$\pm$0.03 & 0.98$\pm$0.01 & 1.00$\pm$0.00 \\
t=2 & 0.84$\pm$0.02 & 0.99$\pm$0.01 & 1.00$\pm$0.00 & 0.84$\pm$0.02 & 0.99$\pm$0.01 & 1.00$\pm$0.00 & 0.84$\pm$0.02 & 0.99$\pm$0.01 & 1.00$\pm$0.00 & 0.84$\pm$0.02 & 0.99$\pm$0.01 & 1.00$\pm$0.00 \\
t=3 & 0.89$\pm$0.03 & 1.00$\pm$0.00 & 1.00$\pm$0.00 & 0.89$\pm$0.03 & 1.00$\pm$0.00 & 1.00$\pm$0.00 & 0.90$\pm$0.02 & 1.00$\pm$0.00 & 1.00$\pm$0.00 & 0.89$\pm$0.03 & 1.00$\pm$0.00 & 1.00$\pm$0.00 \\
\bottomrule
\end{tabular}
}
\end{table*}

\subsection{Settings}
\subsubsection{Dataset.}
Experiments are conducted on four benchmark datasets, including two social networks and two citation networks. BlogCatalog~\cite{LiHTL15} and Flickr~\cite{LiHTL15} are social networks whose node attributes are the posted keywords or tags. Cora~\cite{SenNBGGE08} and Pubmed~\cite{SenNBGGE08} are citation networks in which nodes and edges correspond to documents and citation links. Dataset statistics are summarized in Table~\ref{table:data}. For each dataset, we randomly split the nodes into training (10\%), validation (10\%), and testing (80\%) set. The validation set is employed to tune hyper-parameters and the performance comparison is conducted on the testing set. We report the accuracy of the 2-layer GCN (i.e., our target model) in Table~\ref{table:data}, which is averaged over 5 different random splits.

\subsubsection{Evaluation Protocol.} To evaluate the attack performance of different backdoors, two metrics are employed. The first is the success rate which is the percentage of nodes that changed to the target label after activating the backdoor. Following the previous work~\cite{DBLP:conf/kdd/ZugnerG19}, we display the average of nodes' smallest $q$ that can be successfully backdoored. This metric is referred as avgmin $q$. The smaller such a metric indicates the attack model is better.

\subsubsection{Hyper-parameter Setting.}
We implement all backdoors based on TensorFlow~\cite{DBLP:conf/osdi/AbadiBCCDDDGIIK16}. The target model is set to the 2-layer GCN which is one of the most widely used GCNs~\cite{DBLP:conf/iclr/KipfW17}. Note that there are no hyper-parameters for LGCB and FGSM. The steps $m$ of IG is searched from 10 to 100. We utilize the entire training set as the sampled set of gradient-based backdoors. Following the original work~\cite{DBLP:conf/ijcai/XuC0CWHL19}, PGD iteration steps $T=100$, learning rate $\eta_p=200/\sqrt{p}$, random sampling steps $K=20$. All experiments are run on NVIDIA TITAN RTX.

% $. 

\begin{figure*}[t]
  \centering
  \includegraphics[width=0.24\textwidth]{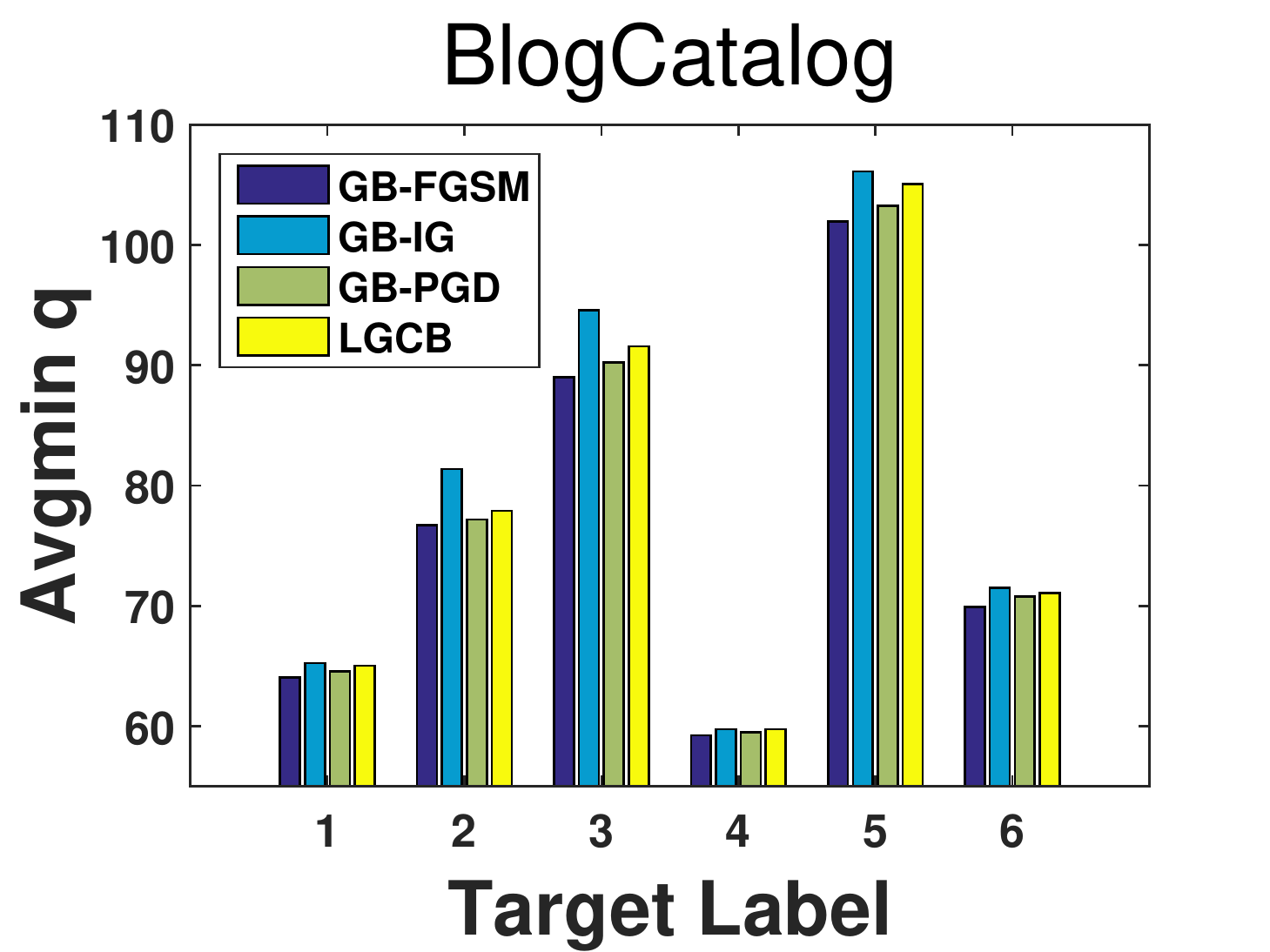}
  \includegraphics[width=0.24\textwidth]{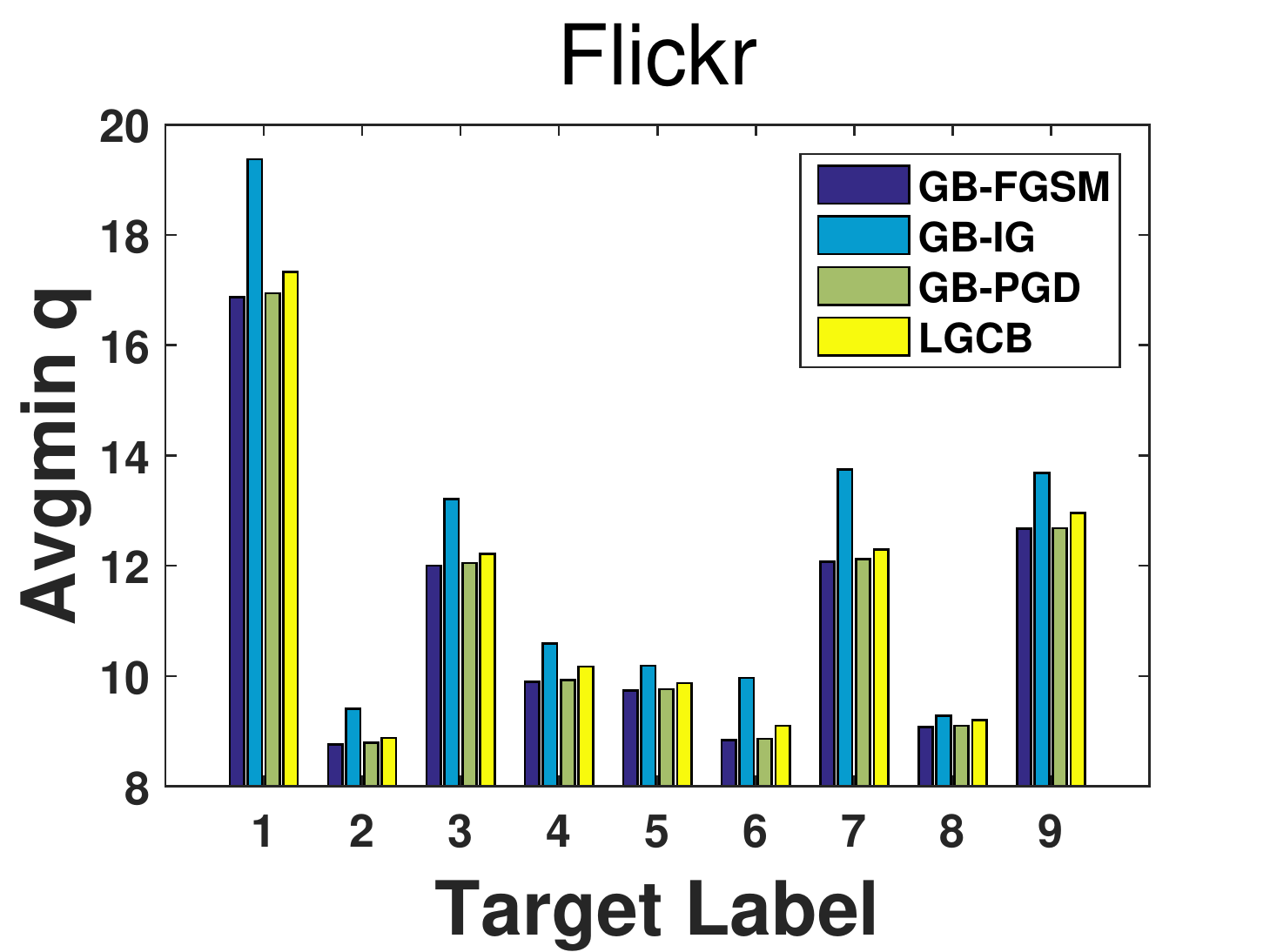}
  \includegraphics[width=0.24\textwidth]{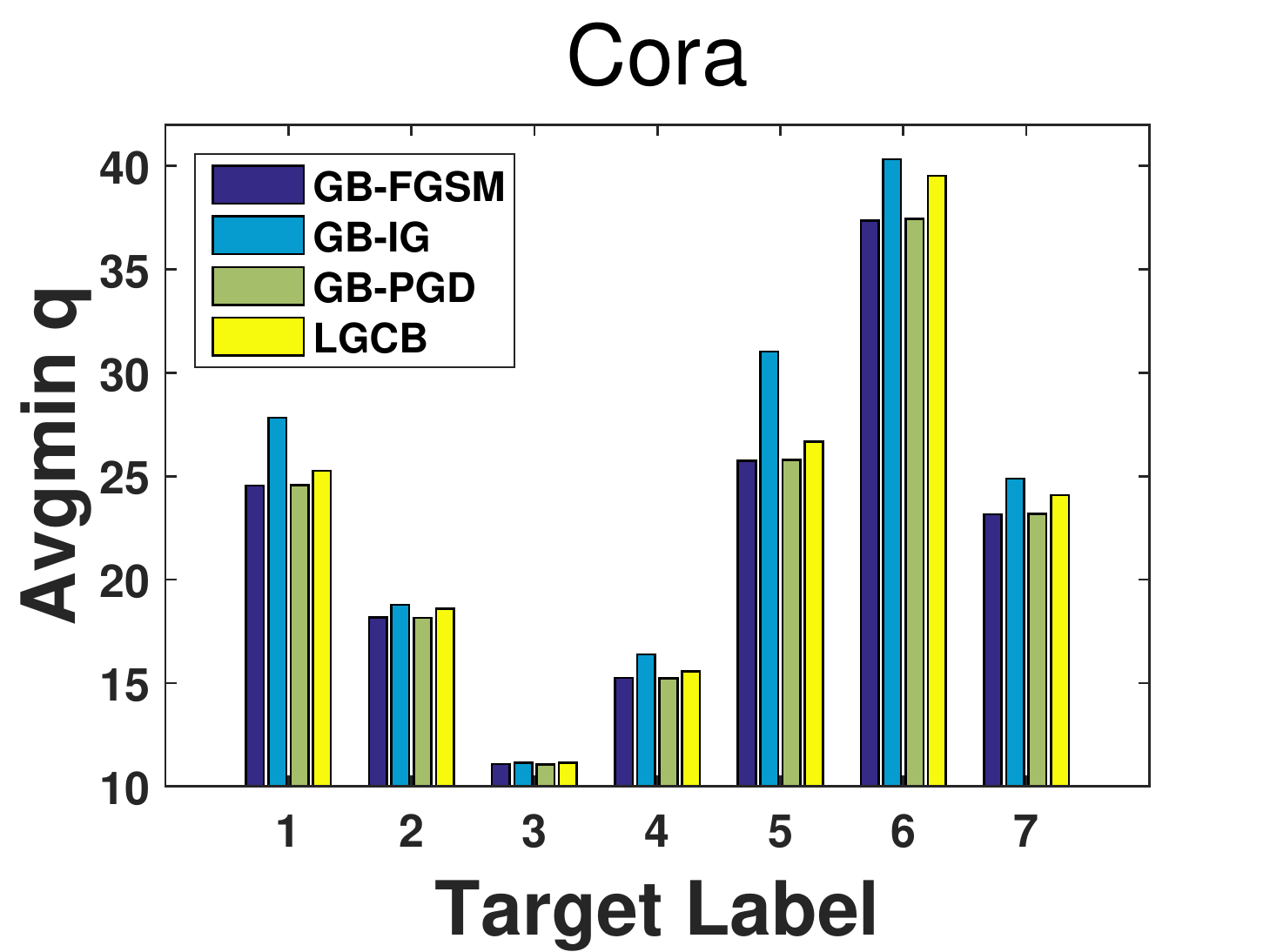}
  \includegraphics[width=0.24\textwidth]{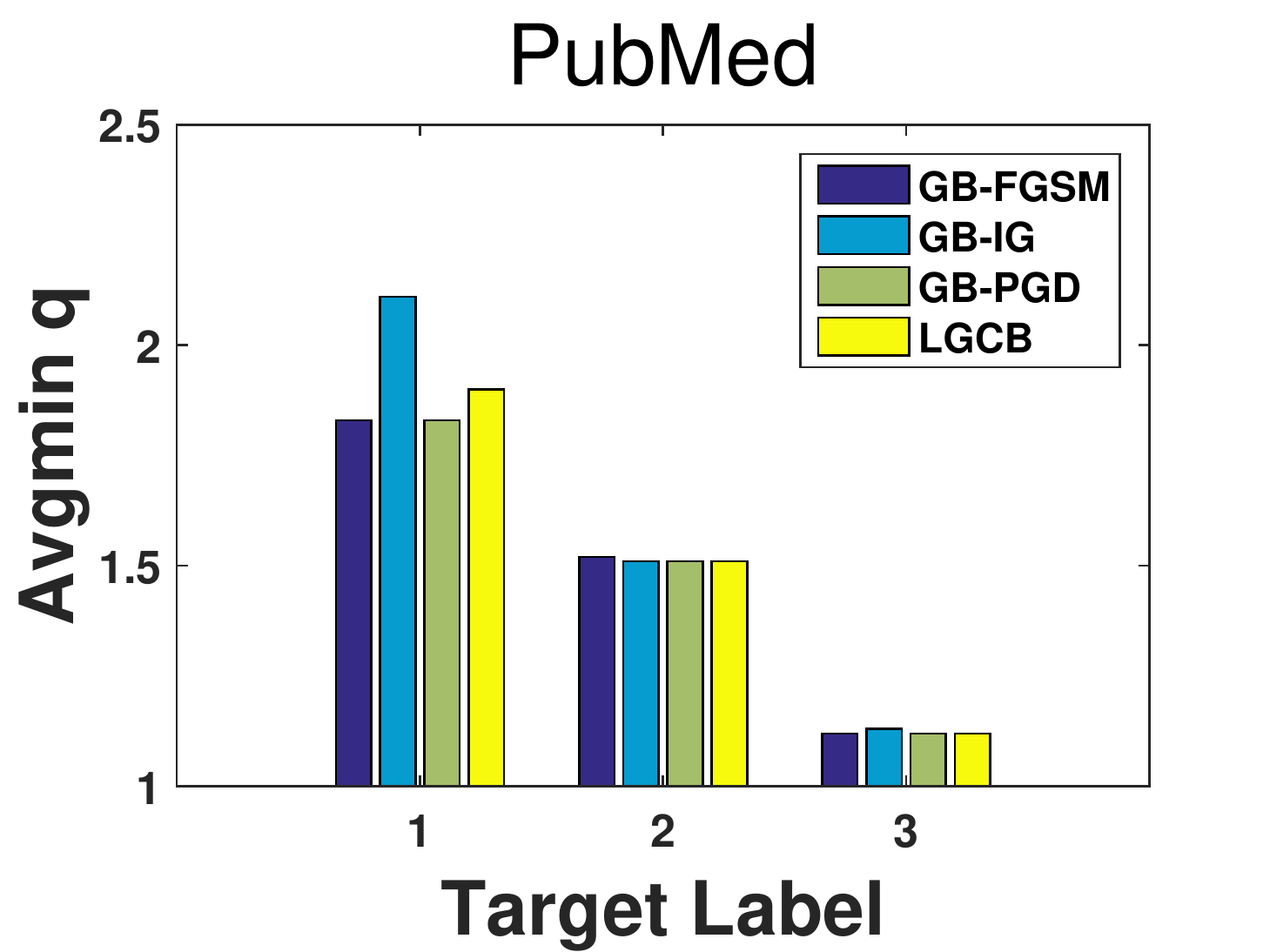}
  \caption{The avgmin $q$ w.r.t. target labels of all backdoors. The results are averaged over 5 runs.}\label{fig:avgmin}
\end{figure*}

\begin{table*}
\small
\caption{Runtime Comparison w.r.t. $q$. The results are averaged over 5 runs. LGCB is significantly faster than gradient-based backdoors. The experiments are run on NVIDIA TITAN RTX.} \label{table:time}
\setlength{\tabcolsep}{4.0mm}
\centering
% \begin{tabular}{@{}c|p{2cm}<{\centering}|p{2cm}<{\centering}|p{2cm}<{\centering}|p{2cm}<{\centering}@{}}
\resizebox{\textwidth}{!}{  %  调整表格大小和页面一致
\begin{tabular}{@{}p{1.5cm}<{\centering}p{0.95cm}<{\centering}p{0.95cm}<{\centering}p{0.95cm}<{\centering}p{0.95cm}<{\centering}p{0.95cm}<{\centering}p{0.95cm}<{\centering}p{0.95cm}<{\centering}p{0.95cm}<{\centering}p{0.95cm}<{\centering}p{0.95cm}<{\centering}p{0.95cm}<{\centering}p{1.2cm}<{\centering}@{}}
\toprule
\multirow{2}{*}{\textbf{Method}}  & \multicolumn{3}{c}{\textbf{BlogCatalog}} & \multicolumn{3}{c}{\textbf{Flickr}} & \multicolumn{3}{c}{\textbf{Cora}} & \multicolumn{3}{c}{\textbf{PubMed}}\\ 
\cmidrule{2-13}
& q=200 & q=300 & q=400  & q=50 & q=100 & q=200 & q=50 & q=100 & q=200 & q=1 & q=10 & q=30\\
\midrule
GB-FGSM & 5.95$\pm$0.02 & 8.89$\pm$0.02 & 11.85$\pm$0.03 & 3.02$\pm$0.01 & 5.95$\pm$0.02 & 11.79$\pm$0.04 & 0.21$\pm$0.00 & 0.41$\pm$0.00 & 0.81$\pm$0.00 & 0.03$\pm$0.01 & 0.11$\pm$0.00 & 0.28$\pm$0.00 \\
GB-IG & 0.65$\pm$0.00 & 0.65$\pm$0.00 & 0.65$\pm$0.00 & 1.29$\pm$0.01 & 1.29$\pm$0.00 & 1.29$\pm$0.01 & 0.07$\pm$0.00 & 0.07$\pm$0.00 & 0.07$\pm$0.00 & 0.18$\pm$0.00 & 0.18$\pm$0.00 & 0.18$\pm$0.00 \\
GB-PGD & 6.03$\pm$0.04 & 6.03$\pm$0.05 & 6.04$\pm$0.10 & 9.46$\pm$0.1 & 9.27$\pm$0.1 & 9.36$\pm$0.2 & 3.22$\pm$0.04 & 3.21$\pm$0.04 & 3.21$\pm$0.02 & 3.91$\pm$0.03 & 3.95$\pm$0.03 & 3.80$\pm$0.02 \\
LGCB & $10^{-3}$ & $10^{-3}$ & $10^{-3}$ & $10^{-3}$ & $10^{-3}$ & $10^{-3}$  &  $10^{-3}$ & $10^{-3}$ & $10^{-3}$  &  $10^{-3}$ & $10^{-3}$ & $10^{-3}$ \\
\bottomrule
\end{tabular}
}
\end{table*}

\subsection{Overall Comparison}

\subsubsection{Overall comparison w.r.t. success rate} The attack success rate of different backdoors w.r.t. different target labels and budget $q$ are reported on Table~\ref{table:success}. Based on the results, we have the following observations:
\begin{itemize}
    \item  Overall, both gradient-based methods (GB-FGSM, GB-IG, and GB-PGD) and LGCB achieve high attack effectiveness. For example, in PubMed dataset, the attack success rates of all backdoors are almost 100\% and reach 100\% when $q=10$ and $q=30$, respectively. These results indicate that GCNs are vulnerable to backdoor attacks. By carefully design the features of trigger nodes, attackers can specify nodes to the target labels. Note that these attacks do not affect the model accuracy on the original tasks unless the triggers are activated which makes them hard to detect.
    \item For gradient-based methods, GB-FGSM and GB-PGD achieve compatible performance and they outperform GB-IG, demonstrating the iterative gradient methods are able to find better triggers. Although LGCB does not utilize the graph data, it achieves better performance than GB-IG in most cases, demonstrating its effectiveness.
    \item From Table~\ref{table:success}, we can infer that a larger $q$ within a certain range indicates a stronger attack performance. However, the number of successfully attacked nodes first increases rapidly, then becomes slowly. For example, when the target label is 1 and the dataset is BlogCatalog, the attack success rate of GB-FGSM only enhances about 6\% when $q$ increases from 200 to 400.
    \item Under the same conditions, some classes are more difficult to be backdoored. For instances, when $q=50$ in Flickr dataset, GB-FGSM achieves 97\% success rate for $t=5$ while only 90\% for $t=0$.
\end{itemize}

\subsubsection{Overall comparison w.r.t. avgmin $q$} To provide a holistic view of the backdoors' performance, the avgmin $q$ of all datasets are displayed in Figure~\ref{fig:avgmin}. We find that:
\begin{itemize}
    \item In most cases, the avgmin q of GB-IG is the largest, indicating it requires more budget to achieve the same attack performance compared to other backdoors.
    \item As can be seen from the figure, in most cases, LGCB only slightly underperforms GB-FGSM and GB-PGD.
    \item It is obvious that some classes are easier to inject backdoor since their avgmin $q$ are much smaller such as the target label 3 in Cora and PubMed datasets.
\end{itemize}

\begin{figure*}[t]
  \centering
  \includegraphics[width=0.24\textwidth]{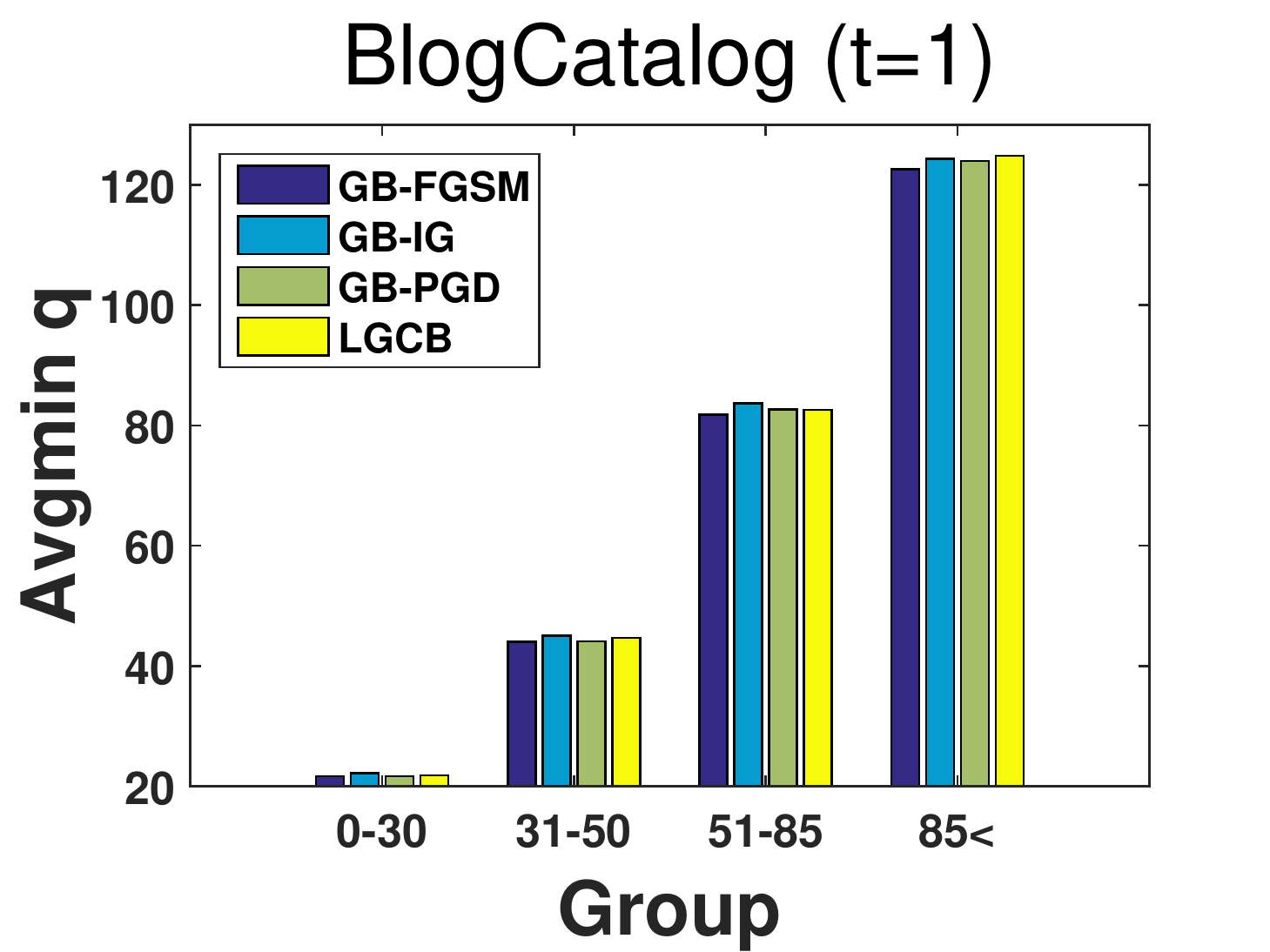}
  \includegraphics[width=0.24\textwidth]{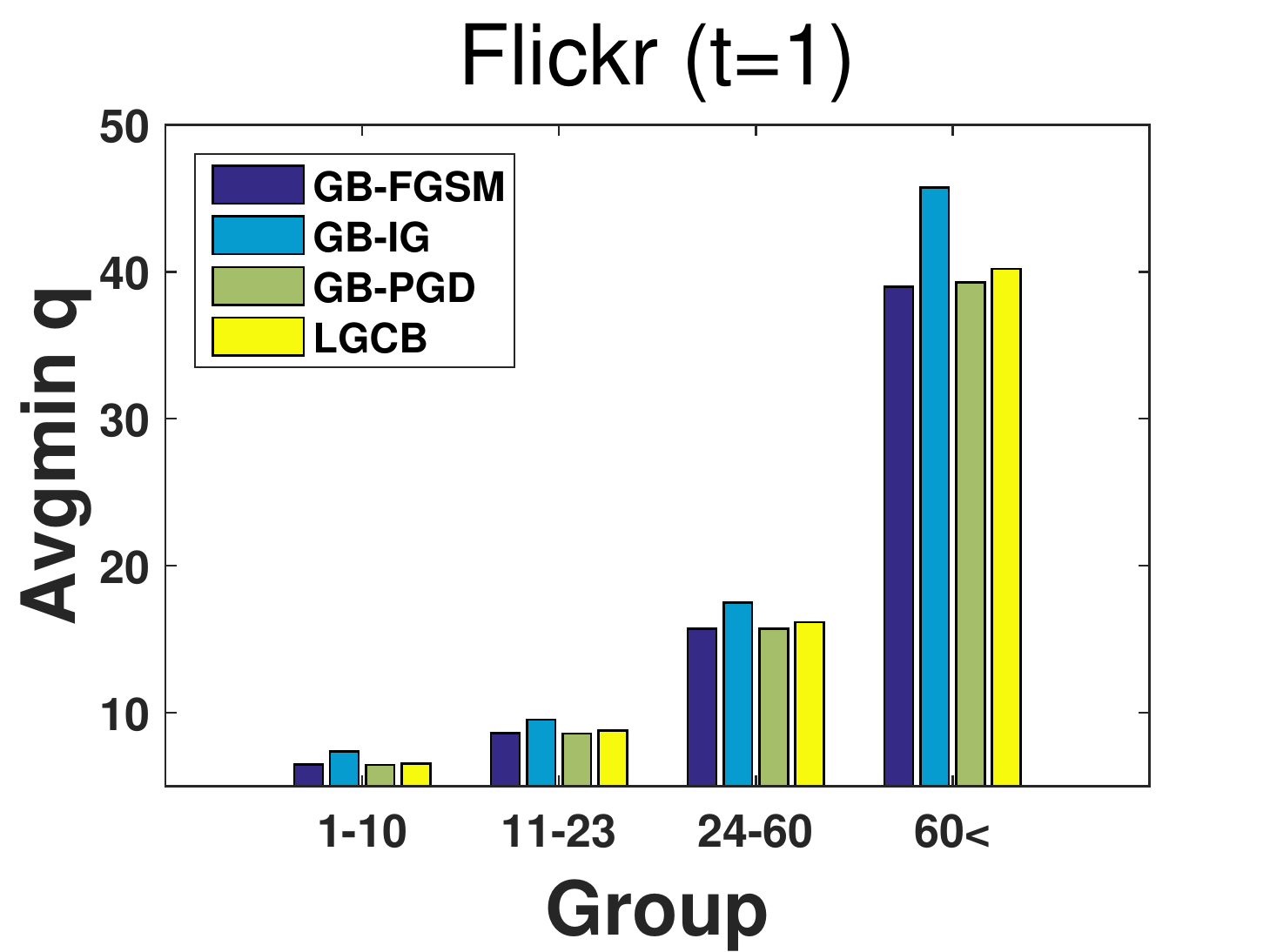}
  \includegraphics[width=0.24\textwidth]{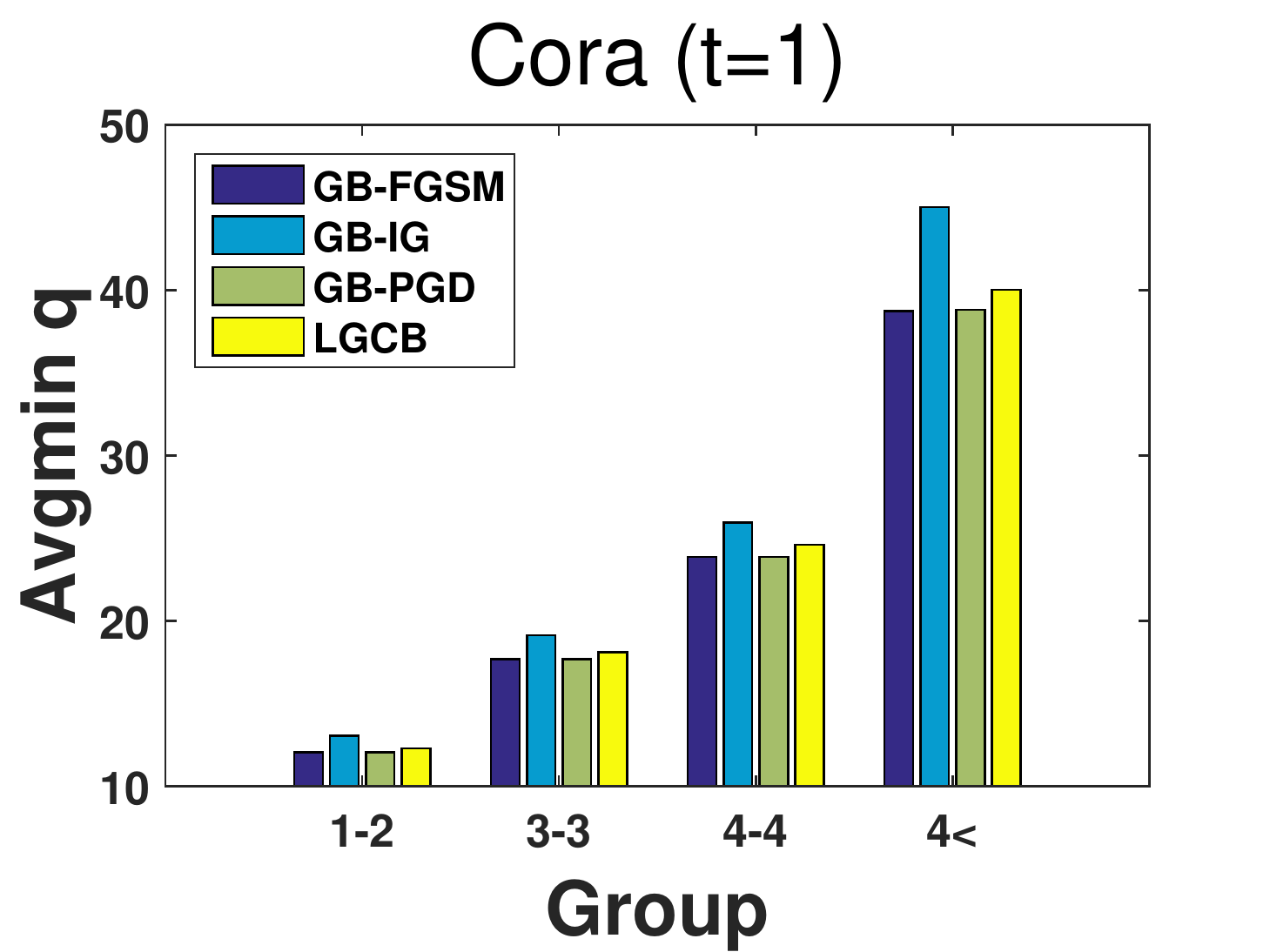}   
  \includegraphics[width=0.24\textwidth]{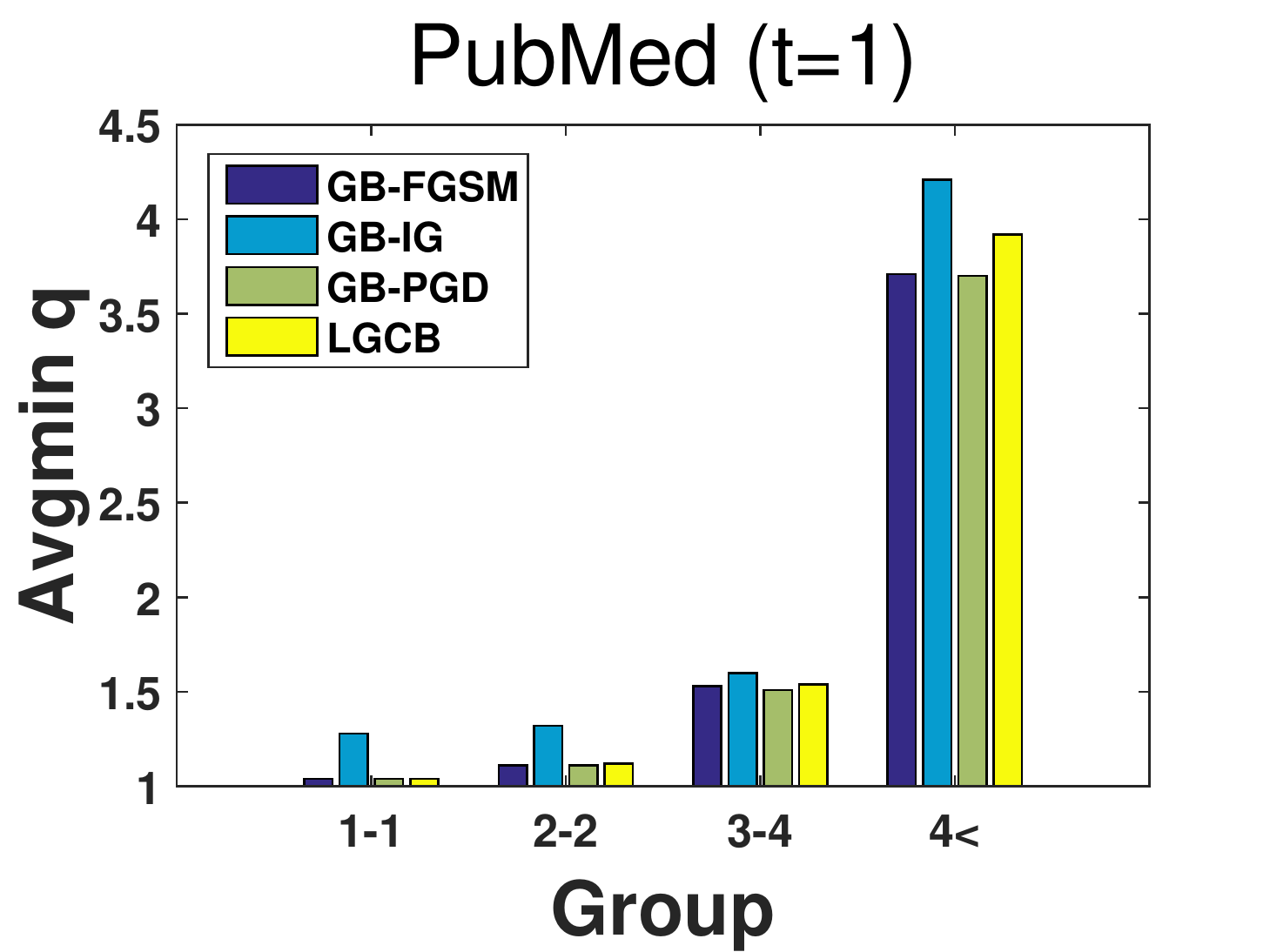}
  \includegraphics[width=0.24\textwidth]{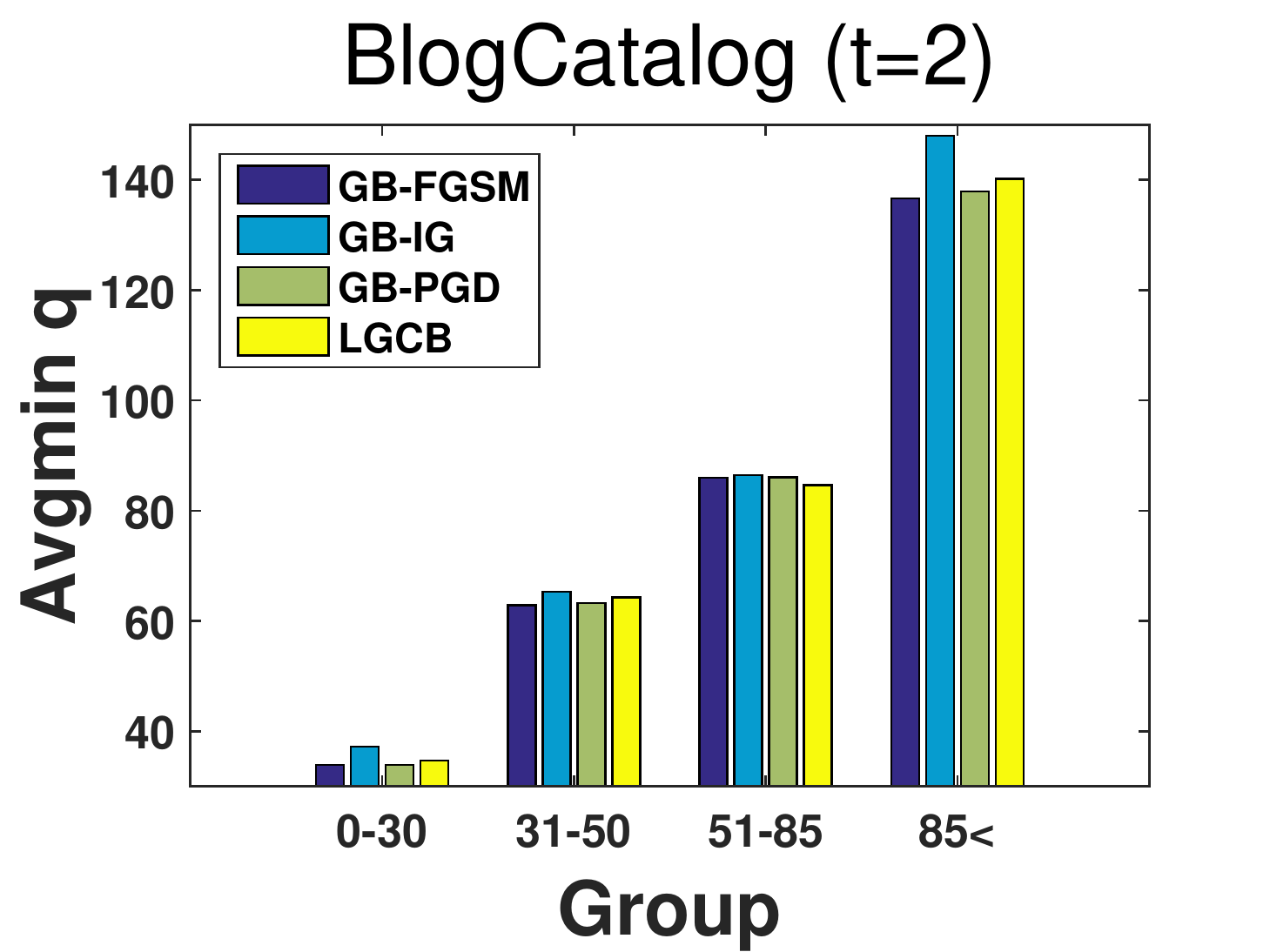}
  \includegraphics[width=0.24\textwidth]{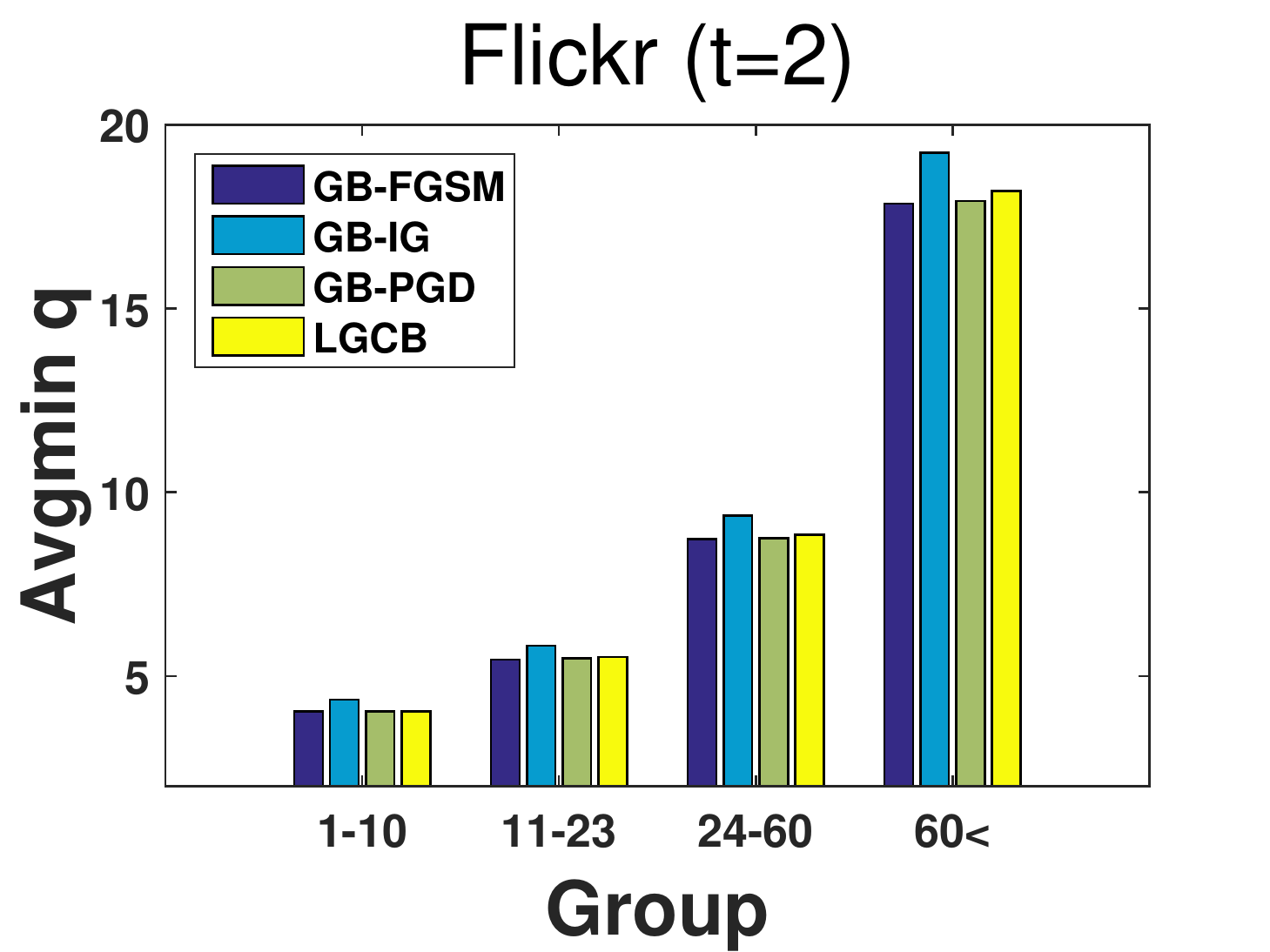}
  \includegraphics[width=0.24\textwidth]{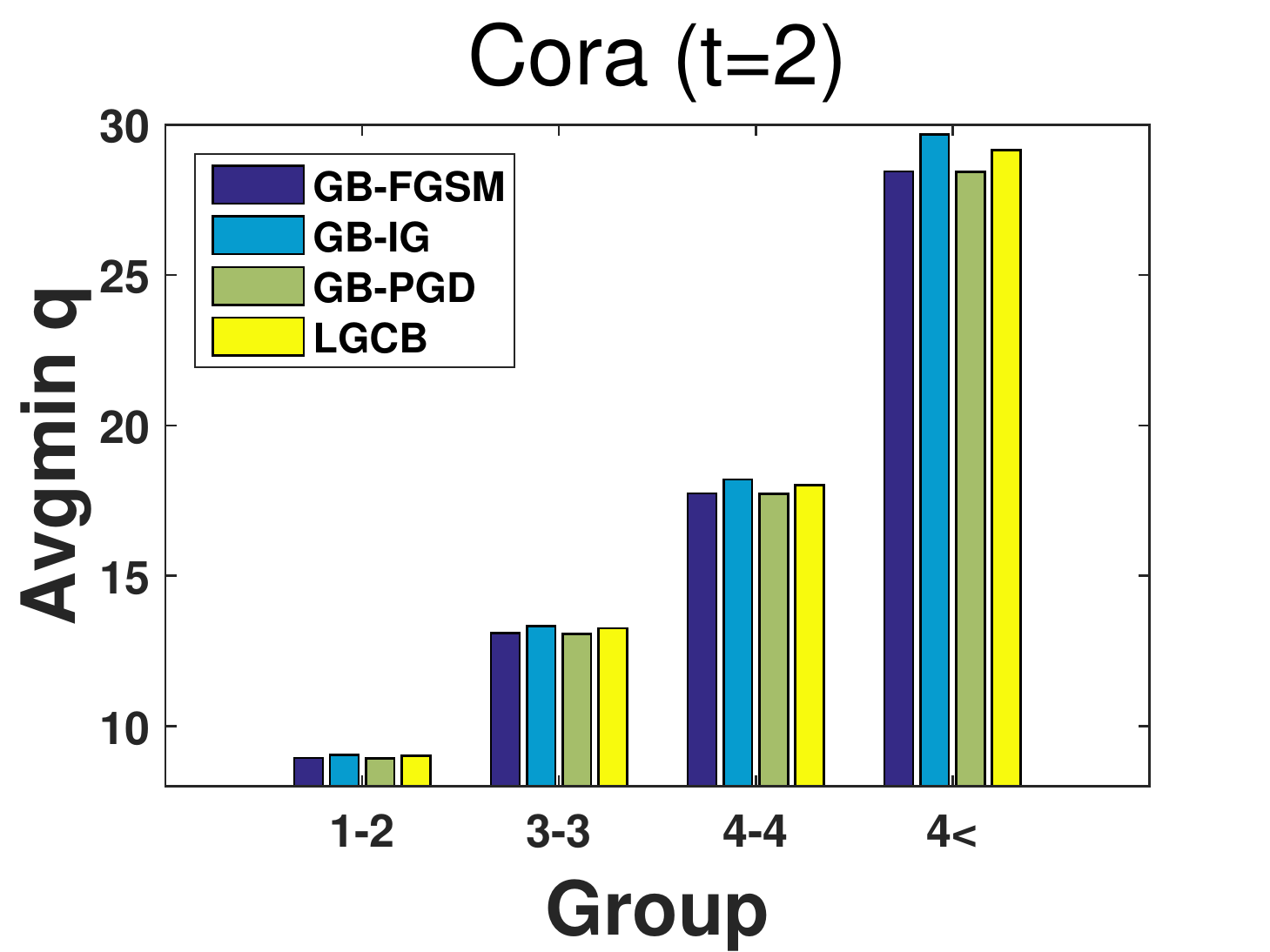}
  \includegraphics[width=0.24\textwidth]{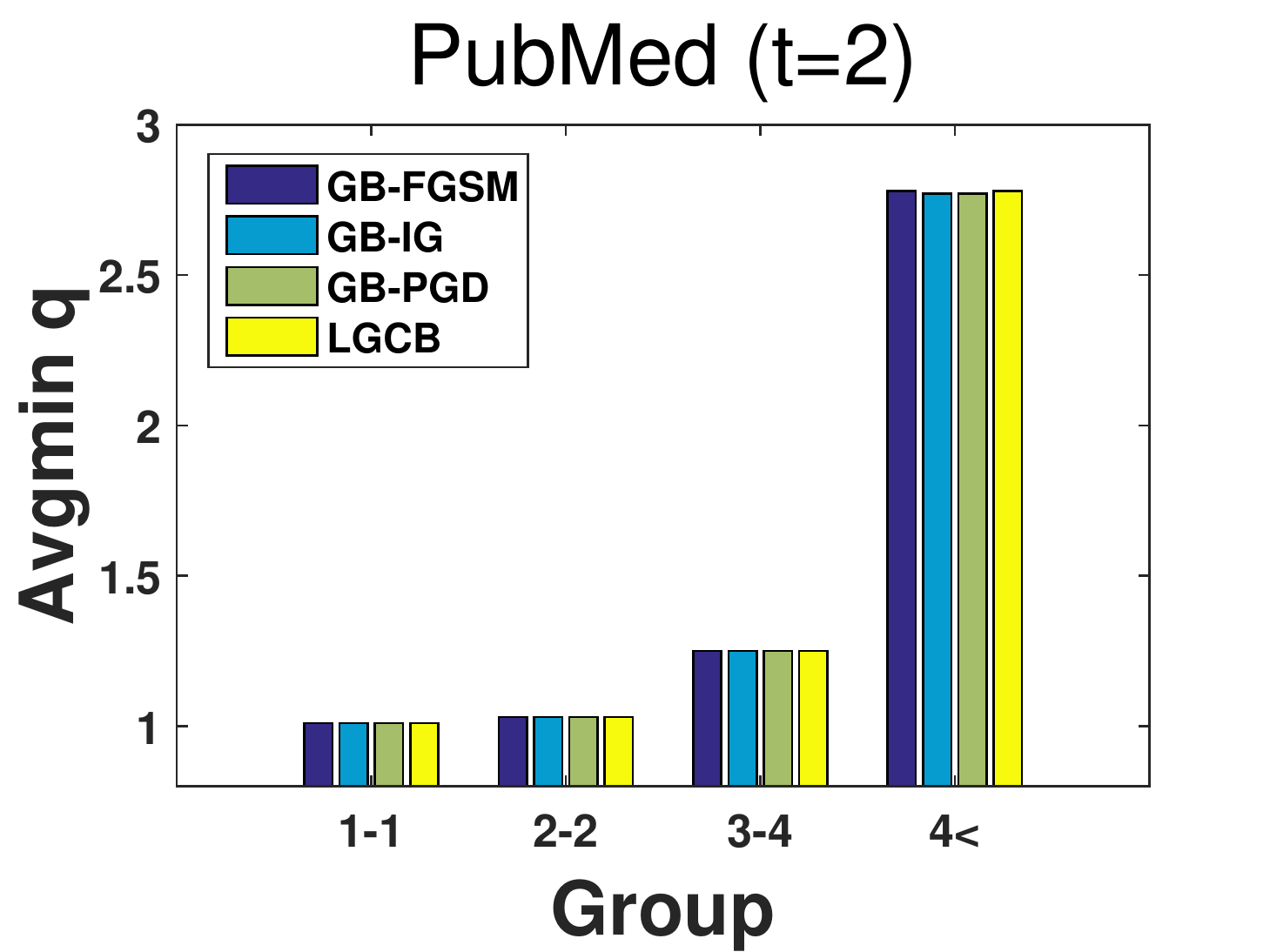}
  \caption{Performance Comparison w.r.t. degree. The node with a larger degree is more robust to backdoor attacks.}\label{fig:degree}
\end{figure*}

\subsection{Runtime Comparison}
To investigate the efficiency of the proposed backdoors, their runtime are displayed in Table~\ref{table:time}. We can observe that:
\begin{itemize}
    \item Increasing $q$ requires more time for GB-FGSM to find a solution while has little impact on GB-IG, GB-PGD, and LGCB. The reason is that GB-FGSM needs to iteratively search the solution according to $q$.
    \item  Under most circumstances, GB-PGD spends more time than GB-FGSM unless $q$ is large such as $q=400$ in BlogCatalog or $q=200$ in Flickr datasets. In summary, the runtime of the four backdoors can be ranked as GB-PGD $>$ GB-FGSM $\gg$ GB-IG $>$ LGCB. 
    \item Although LGCB reaches the compatible performance of GB-FGSM and GB-PGD,  its runtime of LGCB is almost 0 and not relevant to $q$. The reason is that LGCB does not utilize graph data. Thus its runtime is only related to the number of model parameters which is always small and fixed.
\end{itemize}

\subsection{Performance Comparison w.r.t. Node Degree}
Previous work~\cite{DBLP:conf/kdd/ZugnerAG18} has shown that high degree nodes are difficult to be attacked. We want to investigate the effect of node degrees against backdoor attacks. The nodes are divided into four groups. Each group has the same total nodes. We report the avgmin $q$ of all groups w.r.t. target label is 1 and 2 in Table~\ref{fig:degree}. We omit the results of other target labels which have a similar trend. Overall, we can find that nodes with a high degree have a larger avgmin $q$, indicating they are harder to be backdoored compared to nodes with a low degree. Therefore, a possible solution to increase the robustness of GCNs against backdoor attacks is to increase the degree of target nodes such as adversarial immunization~\cite{DBLP:journals/corr/abs-2007-09647}. Moreover, according to the table, we can observe that the performance comparison in groups is consistent with the rank of overall results. That is, GB-FGSM and GB-IG achieve the smallest and largest avgmin $q$. The performance of GB-PGD and LGCB is compatible. 

\subsection{Unnoticeable Trigger}

\begin{figure*}[t]\label{fig:unnoticeable}
  \centering
  \includegraphics[width=0.24\textwidth]{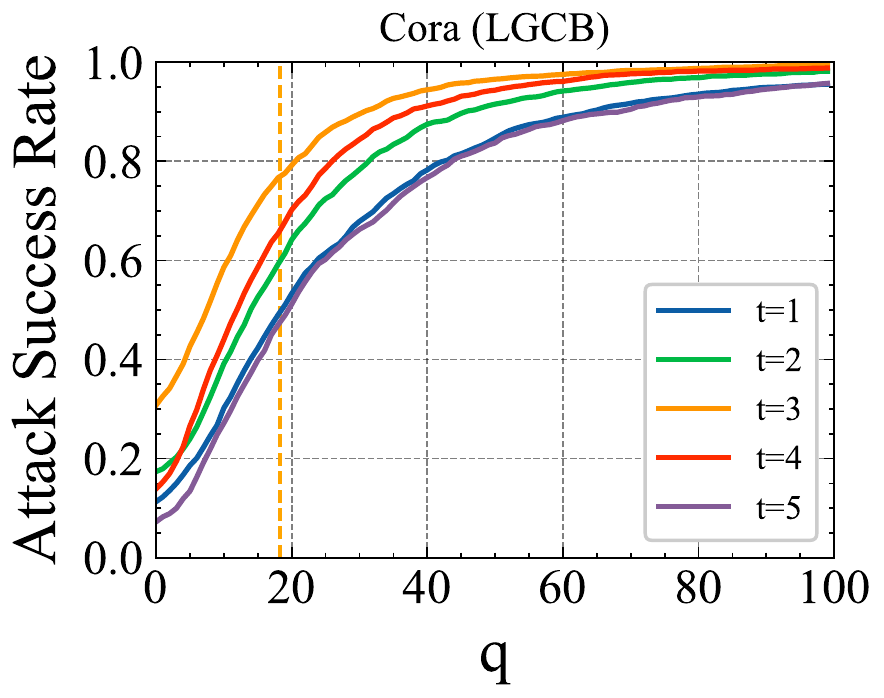}
  \includegraphics[width=0.24\textwidth]{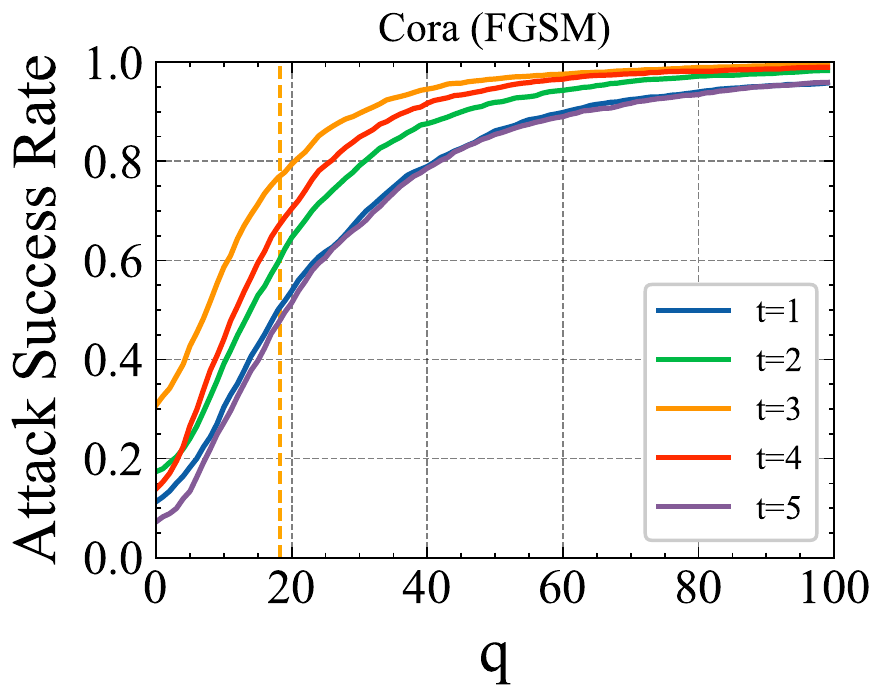}
  \includegraphics[width=0.24\textwidth]{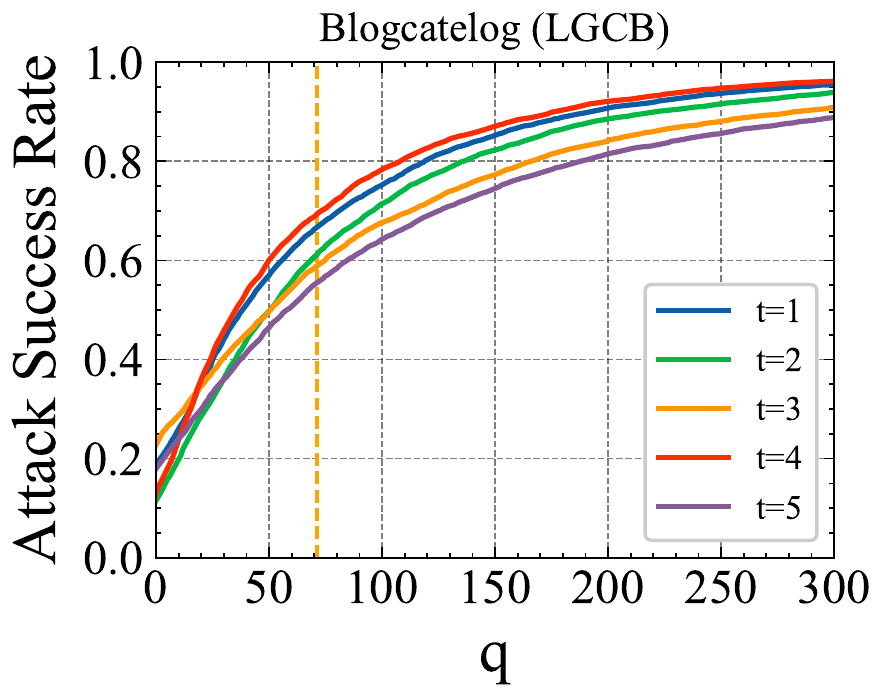}   
  \includegraphics[width=0.24\textwidth]{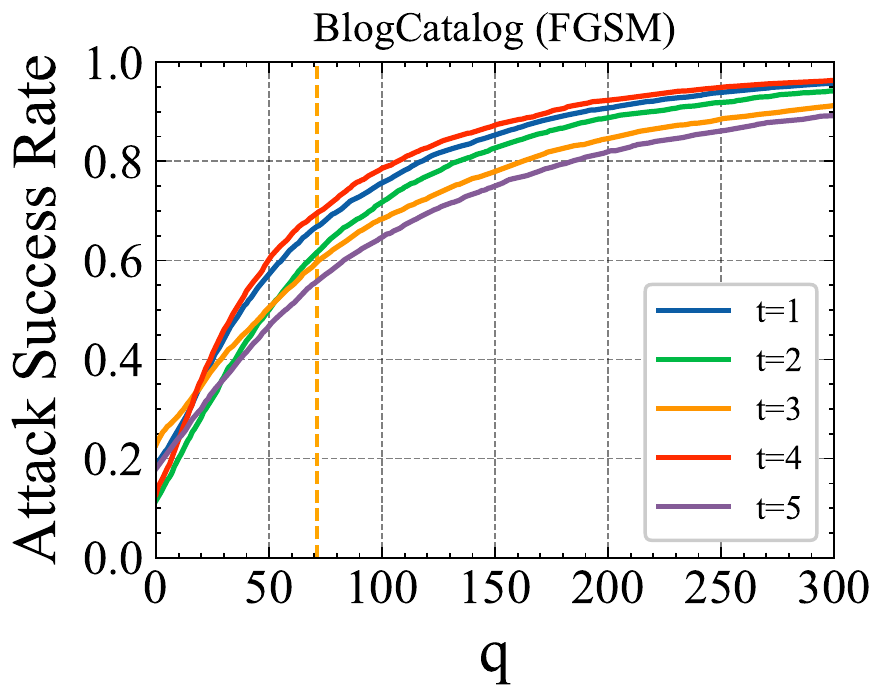}
  \includegraphics[width=0.24\textwidth]{cora_LGCB_sensitivity.pdf}
  \includegraphics[width=0.24\textwidth]{cora_FGSM_sensitivity.pdf}
  \includegraphics[width=0.24\textwidth]{BlogCatalog_LGCB_sensitivity.pdf}   
  \includegraphics[width=0.24\textwidth]{BlogCatalog_LGCB_sensitivity.pdf}
  \caption{Attack success rate w.r.t. attack budget $q$. The orange vertical dashed line denotes the average feature number of the nodes in the graph. The bottom part is tested with unnoticeable generator function while the top part is not.}
\end{figure*}

To investigate whether the proposed triggers can be easily detected, we compare the attack performance and stealthiness with/without unnoticeable generator on different datasets and backdoors which are reported in Fig.~\ref{fig:unnoticeable}:

\begin{itemize}
    \item The attack performance doesn't significantly decrease when the unnoticeable generator function is applied on LGCB and gradient-based backdoors, while the stealthiness is highly improved, which means the proposed unnoticeable trigger is effective and can maintain the stealthiness as well.
    \item It is suffient to backdoor most of the nodes in the graph when the attack budget reaches the average number of the features of these nodes, which means the trigger node is possible to perform as a “average” node in the graph that is not easy to be identified by the feature number.
    \item The attack sensitivities are closed between LGCB and gradient-based backdoors at all levels, which means the conventional numerical solution that solved by gradient optimization can be well approximated by our analytical solution in the linear situation that is more concise and efficient.

\end{itemize}

\subsection{Potential Impact of Trigger Connectivities}

In this section, we briefly discuss the potential impact of the trigger connectivities. For the single node trigger, although the default connectivity that only connect to the target node is effective, there still are potential extra connectivities for the trigger node that may enhance the attack performance. In other words, we are discussing how the design of the vector $\bm{c}$ affect the attack performance. Specifically, we connect the trigger only to the target node (the default case) or connect to the neighbors of the target node. The results are reported in Table~\ref{table:connectivity}:

\begin{itemize}
    \item Even though the trigger node is not directly connected to the target node, it is still possible to attack the target node if extra connectivities are designed elaborately.
    \item Due to the multi-layer strategy of the GCN, the trigger node has opportunity to propagate its features to the target node as the second-order or higher order neighborhood. Thus connecting to the neighbors of the target node is also an effective way to enhance the attack performance.
\end{itemize}

Through the experiments, we see the potential impact of the trigger connectivities. However, we don't dive into this topic too much and leave them as future work.

\begin{table*}[]\label{table:connectivity}
\caption{The attack success rate with/without connectivities to neighbors. $t$ represents the target label.}
\centering
\begin{tabular}{cccccccc}
\toprule
\textbf{Dataset} & \textbf{Connectivities} & t=1 & t=2 & t=3 & t=4 & t=5 & t=6 \\
\midrule
\multirow{2}{*}{\textbf{Blogcatelog}} & target & 87.48 & 94.44 & 91.18 & 76.74 & 62.33 & 78.39 \\
 & target+neighbors & 99.47 & 99.95 & 99.83 & 96.53 & 89.81 & 97.94 \\
 \midrule
\multirow{2}{*}{\textbf{Cora}} & target & 59.96 & 49.80 & 49.74 & 60.08 & 46.45 & 58.58 \\
 & target+neighbors & 75.80 & 78.79 & 71.63 & 86.93 & 66.41 & 78.81 \\
 \midrule
\multirow{2}{*}{\textbf{Flickr}} & target & 72.84 & 87.49 & 80.92 & 84.99 & 86.08 & 86.99 \\
 & target+neighbors & 84.29 & 97.32 & 95.41 & 97.06 & 97.30 & 97.74 \\
\bottomrule
\end{tabular}
\end{table*}

\subsection{Black-box Attack}

In this section, we beyond the white-box assumption and adopt experiments on black-box models. Specifically, the attackers don't know the target model's parameters, architectures, and training procedures. Instead, the surrogate models are developed to obtain the triggers, which are training from the same dataset (the only limited knowledge) with the target model while other settings are different (initilizations, data splits, layers). Then the parameters of the surrogate models are exported to generate the triggers and test on the target model. The results of black- and white-box comparison are reported in Table~\ref{table:black}.

\begin{itemize}
    \item Our proposed backdoor attacks are still effective on black-box models, but not as stable as white-box attack. The attack success rates are ranged from 41.62\%-89.35\% on BlogCatalog and 44.06\%-98.59\% on Cora. It shows the target model is vulnerable even when its parameters are not accessible. 
    \item The attack performance is closed between different attack methods, but varied in different target labels and different datasets. This phenomenon is showed in both black- and white-box attacks, which means the backdoor attack performance is affected not only by the attack methods, but also the dataset itself and the feature-label correlations.
\end{itemize}

\begin{table*}
\caption{The attack success rate under black-box scenarios on BlogCatalog and Cora, where the corresponding white-box scores are noted in parentheses. The results are averaged over 5 runs.} \label{table:black}
% \small
\setlength{\tabcolsep}{4mm}
\centering
% \begin{tabular}{@{}c|p{2cm}<{\centering}|p{2cm}<{\centering}|p{2cm}<{\centering}|p{2cm}<{\centering}@{}}
\resizebox{\textwidth}{!}{
\begin{tabular}{@{}p{1.5cm}<{\centering}p{0.95cm}<{\centering}p{0.95cm}<{\centering}p{0.95cm}<{\centering}p{0.95cm}<{\centering}p{0.95cm}<{\centering}p{0.95cm}<{\centering}p{0.95cm}<{\centering}p{0.95cm}<{\centering}p{0.95cm}<{\centering}p{0.95cm}<{\centering}p{0.95cm}<{\centering}p{1.2cm}<{\centering}@{}}
\toprule
\multirow{2}{*}{\textbf{BlogCatalog}}  & \multicolumn{2}{c}{\textbf{GB-FGSM}} & \multicolumn{2}{c}{\textbf{GB-IG}} & \multicolumn{2}{c}{\textbf{GB-PGD}} & \multicolumn{2}{c}{\textbf{LGCB}}\\ 
\cmidrule{2-9}
& q=200 & q=400 & q=200 & q=400 & q=200 & q=400 & q=200 & q=400 \\
\midrule
t=1 & 53.65(95.03) & 75.46(99.17) &  58.59(93.96) &  77.43(99.00) &  55.10(95.00) &  74.67(99.04) &  60.76(94.80) &  79.29(99.15) \\
t=2 & 45.37(94.52) & 65.66(99.02) &  42.89(94.38) &  64.08(99.10) &  44.55(94.36) &  64.72(98.79) &  49.88(94.38) &  65.58(99.02) \\
t=3 & 43.55(88.68) & 59.36(96.90) &  41.62(85.93) &  59.55(95.28) &  43.06(88.65) &  60.22(96.79) &  43.96(88.61) &  61.12(96.90) \\
t=4 & 76.92(94.42) & 89.35(98.90) &  71.30(93.38)  &  86.96(98.42) & 75.40(94.34)  &  87.18(98.56) & 75.21(94.15) &  88.51(98.77) \\
t=5 & 51.81(84.95) & 72.26(95.38) &  51.94(82.04) &  68.59(92.49) &  52.08(84.91) &  71.25(95.19) &  52.58(83.95) &  71.14(94.52) \\
t=6 & 58.77(92.59) & 74.10(98.33) &  52.62(91.47) &  70.05(98.06) &  59.11(92.46)  &  72.98(98.15) & 57.01(92.38) &  73.32(98.34) \\
\midrule
\midrule
\multirow{2}{*}{\textbf{Cora}}  & \multicolumn{2}{c}{\textbf{GB-FGSM}} & \multicolumn{2}{c}{\textbf{GB-IG}} & \multicolumn{2}{c}{\textbf{GB-PGD}} & \multicolumn{2}{c}{\textbf{LGCB}}\\ 
\cmidrule{2-9}
& q=50 & q=100 & q=50 & q=100 & q=50 & q=100 & q=50 & q=100 \\
\midrule
t=1 & 77.91(90.02) & 90.95(97.06) & 76.22(88.77) & 91.55(97.10) & 77.18(89.94) & 90.50(96.74) & 76.66(89.22) & 92.60(97.10) \\ 
t=2 & 90.30(92.88) & 91.55(97.10) & 89.13(92.15) & 90.50(96.74) & 90.78(92.88) & 92.84(97.14) & 92.03(92.56) & 90.95(97.06) \\ 
t=3 & 91.27(97.10) & 98.11(99.36) & 92.64(97.02) & 98.59(99.32) & 92.11(97.06) & 98.59(99.32) & 92.27(97.02) & 98.63(99.32) \\ 
t=4 & 83.46(85.47) & 95.45(96.26) & 84.87(81.73) & 95.33(94.53) & 83.42(85.43) & 95.45(96.26) & 84.71(84.59) & 95.81(95.77) \\ 
t=5 & 71.43(84.23) & 90.50(95.90) & 71.79(80.52) & 88.05(93.60) & 72.92(84.14) & 90.38(95.73) & 70.58(83.14) & 90.91(95.29) \\ 
t=6 & 50.70(64.02) & 74.37(86.00) & 44.06(52.92) & 65.47(68.81) & 54.08(64.23) & 73.48(85.92) & 44.87(59.40) & 71.23(83.38) \\ 
t=7 & 69.74(91.35) & 87.65(98.15) & 65.11(89.05) & 85.59(97.59) & 69.86(91.31) & 86.12(98.15) & 69.66(91.03) & 85.71(98.15) \\ 
\bottomrule
\end{tabular}
}
\end{table*}

\section{Related Work}
In this work, we review existing work on graph adversarial attacks and backdoor learning which are most relevant to our work.

\subsection{Graph adversarial attacks.}
With the wide applicability of graph neural networks~\cite{DBLP:conf/iclr/KipfW17}, graph adversarial attacks~\cite{DBLP:conf/kdd/ZugnerAG18,DBLP:conf/icml/DaiLTHWZS18,DBLP:conf/iclr/ZugnerG19,DBLP:conf/ijcai/Wu0TDLZ19,DBLP:conf/ijcai/XuC0CWHL19,DBLP:conf/aaai/ChangRXHZC0H20,DBLP:conf/www/LiZHRCH20} have raised increasing attention from researchers. 
The potential attacked tasks ranged from node and graph classification~\cite{DBLP:conf/kdd/ZugnerAG18,DBLP:conf/icml/DaiLTHWZS18} to community detection~\cite{DBLP:conf/www/LiZHRCH20} and malware detection~\cite{DBLP:conf/cikm/HouFZYLWWXS19}. They aim to maximize the accuracy drop on GNN models by modifying the graph structure (e.g., adding malicious edges or nodes) or changing node features. The goal can either be reducing the classification accuracy of a specific target node or the overall accuracy of node classification across the graph. 
For example, NETTACK~\cite{DBLP:conf/kdd/ZugnerAG18} perturb the target node's graph data via a surrogate linear model during the training time and RL-S2V~\cite{DBLP:conf/icml/DaiLTHWZS18} leveraged reinforcement learning techniques to perform attacks on both graph and node classification tasks. Metattack~\cite{DBLP:conf/iclr/ZugnerG19} poisons the whole graph via meta-learning. 

Different from all of the preceding adversarial attacks on graphs, which aim to degrade the model’s generalization accuracy by changing edges or features of the existing nodes, we consider a more limited and practical attack scenario that (1) misclassify arbitrary target node in the graph (2) to target label (3) by a single node injection. Several previous works~\cite{2018/arXiv/AddingFakeNodes,2019/bigdataconf/Takahashi19,2020/datamine/WangLSLYZ20,2020/www/NIPA_SunWTHH20,2021/CIKM/G-NIA} have studied the topic in similar attack scenario. All of these attacks are achieved by adding nodes rather than edge or feature perturbations on the existing nodes, which is more practical since the attackers don't need access to the target node or other normal nodes in the graph. \cite{2018/arXiv/AddingFakeNodes} extend the greedy optimization approach of Nettack~\cite{DBLP:conf/kdd/ZugnerAG18} to iteratively generating the features of fake nodes, thus suffering from the inefficient and sub-optimal problem. AFGSM~\cite{2020/www/NIPA_SunWTHH20} proposes an approximated solution to generate features of injected nodes efficiently. G-NIA~\cite{2021/CIKM/G-NIA} propose a generalizable node injection attack model. However, most of them simply extend the gradient optimization based approaches which are not efficient enough and leave the information in the trained parametic models unexplored. Our proposed LGCB can get rid of repeated gradient computation and efficiently generated the trigger that can universally adapt to almost all nodes in the graph at inference phase.
 
\subsection{Backdoor learning.}
Multiple papers~\cite{DBLP:journals/access/GuLDG19,DBLP:conf/cvpr/RakinHF20,DBLP:conf/kdd/TangDLYH20,DBLP:conf/ccs/YaoLZZ19,DBLP:conf/nips/Tran0M18} have revealed that deep neural networks in the image domain are vulnerable to backdoor attacks. BadNets~\cite{DBLP:journals/corr/abs-1708-06733} is firstly proposed to fine-tune the pre-trained model on a poisoned dataset which is constructed by injecting a backdoor trigger (e.g., a patch) to a certain proportion of training data. However, this approach need explicitly change both the input (i.e., inject triggers) and output (i.e., target label), which might be easily discovered by the users. 
As such several non-poisoning backdoors ~\cite{DBLP:conf/cvpr/RakinHF20,DBLP:conf/kdd/TangDLYH20,DBLP:journals/corr/abs-1812-03128} are proposed to implicitly inject triggers. 
Recent works have further studies backdoor attacks on other tasks such as natural language processing~\cite{DBLP:journals/access/DaiCL19,DBLP:conf/acl/KuritaMN20} and reinforcement learning~\cite{DBLP:journals/corr/abs-1903-06638}. 
Traditional backdoors in the image domain utilize a specific pattern (e.g., stickers or dots) as the trigger and need to modified the model such as retraining the model or parameters perturbation. 
In parallel to our work, there are two works~\cite{DBLP:journals/corr/abs-2006-11890,DBLP:journals/corr/abs-2006-11165} that studied the backdoor attacks on graph neural networks as well. However, they need to modify the model parameters, and the model accuracy drops. 
Our work proposes a novel backdoor on graph convolutional networks whose trigger is set as a single node and not need to change parameters of the target model.

\section{Conclusion and Future Work}
This work proposes neighboring backdoors on graph convolutional networks. By specifically designing the features of the single node trigger, we can inject backdoors into multiple classes simultaneously without harming the model performance on the original task. Unlike the previous backdoors on the image classification task, our methods do not require retraining the target model on a poisoning set. Two types of backdoors are proposed to generate the trigger features. Extensive experiments on four datasets demonstrate that all proposed backdoors achieve high attack success rates on GCNs. 

In the future, we plan to extend our work in the following three directions: (1) besides GCNs, there are many variants of graph neural networks such as GraphSAGE~\cite{DBLP:conf/nips/HamiltonYL17} and GAT~\cite{DBLP:conf/iclr/VelickovicCCRLB18}. We will study whether these GNNs are vulnerable to backdoor attacks; (2) improve the robustness of GCNs against backdoor attacks is another important task. We will investigate whether existing robust GCNs can defend our proposed graph backdoors. 
% (3) explore the attack methods for more generalized triggers such as multiple node triggers and connective triggers.

% \begin{acks}
% \end{acks}

% \cite{*}
\bibliographystyle{IEEEtran}
\bibliography{sample-base}

\begin{thebibliography}{10}
\providecommand{\url}[1]{#1}
\csname url@samestyle\endcsname
\providecommand{\newblock}{\relax}
\providecommand{\bibinfo}[2]{#2}
\providecommand{\BIBentrySTDinterwordspacing}{\spaceskip=0pt\relax}
\providecommand{\BIBentryALTinterwordstretchfactor}{4}
\providecommand{\BIBentryALTinterwordspacing}{\spaceskip=\fontdimen2\font plus
\BIBentryALTinterwordstretchfactor\fontdimen3\font minus
  \fontdimen4\font\relax}
\providecommand{\BIBforeignlanguage}[2]{{%
\expandafter\ifx\csname l@#1\endcsname\relax
\typeout{** WARNING: IEEEtran.bst: No hyphenation pattern has been}%
\typeout{** loaded for the language `#1'. Using the pattern for}%
\typeout{** the default language instead.}%
\else
\language=\csname l@#1\endcsname
\fi
#2}}
\providecommand{\BIBdecl}{\relax}
\BIBdecl

\bibitem{DBLP:conf/iclr/KipfW17}
T.~N. Kipf and M.~Welling, ``Semi-supervised classification with graph
  convolutional networks,'' in \emph{ICLR}, 2017.

\bibitem{DBLP:conf/sigir/Wang0WFC19}
X.~Wang, X.~He, M.~Wang, F.~Feng, and T.~Chua, ``Neural graph collaborative
  filtering,'' in \emph{SIGIR}, 2019, pp. 165--174.

\bibitem{DBLP:conf/sigir/0001DWLZ020}
X.~He, K.~Deng, X.~Wang, Y.~Li, Y.~Zhang, and M.~Wang, ``Lightgcn: Simplifying
  and powering graph convolution network for recommendation,'' in \emph{SIGIR},
  2020, pp. 639--648.

\bibitem{DBLP:conf/aaai/WuLXWC20}
Y.~Wu, D.~Lian, Y.~Xu, L.~Wu, and E.~Chen, ``Graph convolutional networks with
  markov random field reasoning for social spammer detection,'' in \emph{AAAI},
  2020, pp. 1054--1061.

\bibitem{DBLP:conf/aaai/BianXXZHRH20}
T.~Bian, X.~Xiao, T.~Xu, P.~Zhao, W.~Huang, Y.~Rong, and J.~Huang, ``Rumor
  detection on social media with bi-directional graph convolutional networks,''
  in \emph{AAAI}, 2020, pp. 549--556.

\bibitem{DBLP:conf/cikm/DongZHSL19}
M.~Dong, B.~Zheng, N.~Q.~V. Hung, H.~Su, and G.~Li, ``Multiple rumor source
  detection with graph convolutional networks,'' in \emph{CIKM}, 2019, pp.
  569--578.

\bibitem{DBLP:journals/corr/abs-2007-08745}
\BIBentryALTinterwordspacing
Y.~Li, B.~Wu, Y.~Jiang, Z.~Li, and S.~Xia, ``Backdoor learning: {A} survey,''
  \emph{CoRR}, vol. abs/2007.08745, 2020. [Online]. Available:
  \url{https://arxiv.org/abs/2007.08745}
\BIBentrySTDinterwordspacing

\bibitem{DBLP:conf/kdd/TangDLYH20}
R.~Tang, M.~Du, N.~Liu, F.~Yang, and X.~Hu, ``An embarrassingly simple approach
  for trojan attack in deep neural networks,'' in \emph{KDD}, 2020, pp.
  218--228.

\bibitem{DBLP:conf/kdd/ZugnerAG18}
D.~Z{\"{u}}gner, A.~Akbarnejad, and S.~G{\"{u}}nnemann, ``Adversarial attacks
  on neural networks for graph data,'' in \emph{KDD}, 2018, pp. 2847--2856.

\bibitem{DBLP:conf/icml/DaiLTHWZS18}
H.~Dai, H.~Li, T.~Tian, X.~Huang, L.~Wang, J.~Zhu, and L.~Song, ``Adversarial
  attack on graph structured data,'' in \emph{ICML}, 2018, pp. 1123--1132.

\bibitem{DBLP:conf/iclr/ZugnerG19}
D.~Z{\"{u}}gner and S.~G{\"{u}}nnemann, ``Adversarial attacks on graph neural
  networks via meta learning,'' in \emph{ICLR}, 2019.

\bibitem{DBLP:conf/ijcai/Wu0TDLZ19}
H.~Wu, C.~Wang, Y.~Tyshetskiy, A.~Docherty, K.~Lu, and L.~Zhu, ``Adversarial
  examples for graph data: Deep insights into attack and defense,'' in
  \emph{IJCAI}, 2019, pp. 4816--4823.

\bibitem{DBLP:conf/ijcai/XuC0CWHL19}
K.~Xu, H.~Chen, S.~Liu, P.~Chen, T.~Weng, M.~Hong, and X.~Lin, ``Topology
  attack and defense for graph neural networks: An optimization perspective,''
  in \emph{IJCAI}, 2019, pp. 3961--3967.

\bibitem{DBLP:conf/aaai/ChangRXHZC0H20}
H.~Chang, Y.~Rong, T.~Xu, W.~Huang, H.~Zhang, P.~Cui, W.~Zhu, and J.~Huang, ``A
  restricted black-box adversarial framework towards attacking graph embedding
  models,'' in \emph{AAAI}, 2020, pp. 3389--3396.

\bibitem{DBLP:conf/www/LiZHRCH20}
J.~Li, H.~Zhang, Z.~Han, Y.~Rong, H.~Cheng, and J.~Huang, ``Adversarial attack
  on community detection by hiding individuals,'' in \emph{WWW}, 2020, pp.
  917--927.

\bibitem{DBLP:conf/www/SunWTHH20}
Y.~Sun, S.~Wang, X.~Tang, T.~Hsieh, and V.~G. Honavar, ``Adversarial attacks on
  graph neural networks via node injections: {A} hierarchical reinforcement
  learning approach,'' in \emph{WWW}, 2020, pp. 673--683.

\bibitem{DBLP:journals/corr/abs-2006-11165}
Z.~Zhang, J.~Jia, B.~Wang, and N.~Z. Gong, ``Backdoor attacks to graph neural
  networks,'' \emph{CoRR}, vol. abs/2006.11165, 2020.

\bibitem{DBLP:journals/corr/abs-2006-11890}
Z.~Xi, R.~Pang, S.~Ji, and T.~Wang, ``Graph backdoor,'' \emph{CoRR}, vol.
  abs/2006.11890, 2020.

\bibitem{DBLP:conf/kdd/ZugnerG19}
D.~Z{\"{u}}gner and S.~G{\"{u}}nnemann, ``Certifiable robustness and robust
  training for graph convolutional networks,'' in \emph{KDD}, 2019, pp.
  246--256.

\bibitem{DBLP:conf/kdd/ZugnerG20}
------, ``Certifiable robustness of graph convolutional networks under
  structure perturbations,'' in \emph{KDD}, 2020, pp. 1656--1665.

\bibitem{DBLP:conf/nips/BojchevskiG19}
A.~Bojchevski and S.~G{\"{u}}nnemann, ``Certifiable robustness to graph
  perturbations,'' in \emph{NeurIPS}, 2019, pp. 8317--8328.

\bibitem{DBLP:journals/corr/GoodfellowSS14}
I.~J. Goodfellow, J.~Shlens, and C.~Szegedy, ``Explaining and harnessing
  adversarial examples,'' in \emph{ICLR}, 2015.

\bibitem{DBLP:conf/iclr/MadryMSTV18}
A.~Madry, A.~Makelov, L.~Schmidt, D.~Tsipras, and A.~Vladu, ``Towards deep
  learning models resistant to adversarial attacks,'' in \emph{ICLR}, 2018.

\bibitem{LiHTL15}
J.~Li, X.~Hu, J.~Tang, and H.~Liu, ``Unsupervised streaming feature selection
  in social media,'' in \emph{CIKM}, J.~Bailey, A.~Moffat, C.~C. Aggarwal,
  M.~de~Rijke, R.~Kumar, V.~Murdock, T.~K. Sellis, and J.~X. Yu, Eds.\hskip 1em
  plus 0.5em minus 0.4em\relax {ACM}, 2015, pp. 1041--1050.

\bibitem{SenNBGGE08}
P.~Sen, G.~Namata, M.~Bilgic, L.~Getoor, B.~Gallagher, and T.~Eliassi{-}Rad,
  ``Collective classification in network data,'' \emph{{AI} Mag.}, vol.~29,
  no.~3, pp. 93--106, 2008.

\bibitem{DBLP:conf/osdi/AbadiBCCDDDGIIK16}
M.~Abadi, P.~Barham, J.~Chen, Z.~Chen, A.~Davis, J.~Dean, M.~Devin,
  S.~Ghemawat, G.~Irving, M.~Isard, M.~Kudlur, J.~Levenberg, R.~Monga,
  S.~Moore, D.~G. Murray, B.~Steiner, P.~A. Tucker, V.~Vasudevan, P.~Warden,
  M.~Wicke, Y.~Yu, and X.~Zheng, ``Tensorflow: {A} system for large-scale
  machine learning,'' in \emph{OSDI}, 2016, pp. 265--283.

\bibitem{DBLP:journals/corr/abs-2007-09647}
S.~Tao, H.~Shen, Q.~Cao, L.~Hou, and X.~Cheng, ``Adversarial immunization for
  improving certifiable robustness on graphs,'' \emph{CoRR}, vol.
  abs/2007.09647, 2020.

\bibitem{DBLP:conf/cikm/HouFZYLWWXS19}
S.~Hou, Y.~Fan, Y.~Zhang, Y.~Ye, J.~Lei, W.~Wan, J.~Wang, Q.~Xiong, and
  F.~Shao, ``\emph{{\(\alpha\)}Cyber}: Enhancing robustness of android malware
  detection system against adversarial attacks on heterogeneous graph based
  model,'' in \emph{CIKM}, 2019, pp. 609--618.

\bibitem{2018/arXiv/AddingFakeNodes}
X.~Wang, J.~Eaton, C.~Hsieh, and S.~F. Wu, ``Attack graph convolutional
  networks by adding fake nodes,'' \emph{CoRR}, vol. abs/1810.10751, 2018.

\bibitem{2019/bigdataconf/Takahashi19}
T.~Takahashi, ``Indirect adversarial attacks via poisoning neighbors for graph
  convolutional networks,'' in \emph{Bigdata}.\hskip 1em plus 0.5em minus
  0.4em\relax {IEEE}, 2019, pp. 1395--1400.

\bibitem{2020/datamine/WangLSLYZ20}
J.~Wang, M.~Luo, F.~Suya, J.~Li, Z.~Yang, and Q.~Zheng, ``Scalable attack on
  graph data by injecting vicious nodes,'' \emph{Data Min. Knowl. Discov.}, pp.
  1363--1389, 2020.

\bibitem{2020/www/NIPA_SunWTHH20}
Y.~Sun, S.~Wang, X.~Tang, T.~Hsieh, and V.~G. Honavar, ``Adversarial attacks on
  graph neural networks via node injections: {A} hierarchical reinforcement
  learning approach,'' in \emph{WWW}, 2020, pp. 673--683.

\bibitem{2021/CIKM/G-NIA}
S.~Tao, Q.~Cao, H.~Shen, J.~Huang, Y.~Wu, and X.~Cheng, ``Single node injection
  attack against graph neural networks,'' 2021.

\bibitem{DBLP:journals/access/GuLDG19}
T.~Gu, K.~Liu, B.~Dolan{-}Gavitt, and S.~Garg, ``Badnets: Evaluating
  backdooring attacks on deep neural networks,'' \emph{{IEEE} Access}, vol.~7,
  pp. 47\,230--47\,244, 2019.

\bibitem{DBLP:conf/cvpr/RakinHF20}
A.~S. Rakin, Z.~He, and D.~Fan, ``{TBT:} targeted neural network attack with
  bit trojan,'' in \emph{CVPR}, 2020, pp. 13\,195--13\,204.

\bibitem{DBLP:conf/ccs/YaoLZZ19}
Y.~Yao, H.~Li, H.~Zheng, and B.~Y. Zhao, ``Latent backdoor attacks on deep
  neural networks,'' in \emph{CCS}, 2019, pp. 2041--2055.

\bibitem{DBLP:conf/nips/Tran0M18}
B.~Tran, J.~Li, and A.~Madry, ``Spectral signatures in backdoor attacks,'' in
  \emph{NeurIPS}, 2018, pp. 8011--8021.

\bibitem{DBLP:journals/corr/abs-1708-06733}
T.~Gu, B.~Dolan{-}Gavitt, and S.~Garg, ``Badnets: Identifying vulnerabilities
  in the machine learning model supply chain,'' \emph{CoRR}, vol.
  abs/1708.06733, 2017.

\bibitem{DBLP:journals/corr/abs-1812-03128}
J.~Dumford and W.~J. Scheirer, ``Backdooring convolutional neural networks via
  targeted weight perturbations,'' \emph{CoRR}, vol. abs/1812.03128, 2018.

\bibitem{DBLP:journals/access/DaiCL19}
J.~Dai, C.~Chen, and Y.~Li, ``A backdoor attack against lstm-based text
  classification systems,'' \emph{{IEEE} Access}, vol.~7, pp.
  138\,872--138\,878, 2019.

\bibitem{DBLP:conf/acl/KuritaMN20}
K.~Kurita, P.~Michel, and G.~Neubig, ``Weight poisoning attacks on pretrained
  models,'' in \emph{ACL}, 2020, pp. 2793--2806.

\bibitem{DBLP:journals/corr/abs-1903-06638}
P.~Kiourti, K.~Wardega, S.~Jha, and W.~Li, ``Trojdrl: Trojan attacks on deep
  reinforcement learning agents,'' \emph{CoRR}, vol. abs/1903.06638, 2019.

\bibitem{DBLP:conf/nips/HamiltonYL17}
W.~L. Hamilton, Z.~Ying, and J.~Leskovec, ``Inductive representation learning
  on large graphs,'' in \emph{NIPS}, 2017, pp. 1024--1034.

\bibitem{DBLP:conf/iclr/VelickovicCCRLB18}
P.~Velickovic, G.~Cucurull, A.~Casanova, A.~Romero, P.~Li{\`{o}}, and
  Y.~Bengio, ``Graph attention networks,'' in \emph{ICLR}, 2018.

\end{thebibliography}


\begin{IEEEbiography}[{\includegraphics[width=1in,height=1.25in,clip,keepaspectratio]{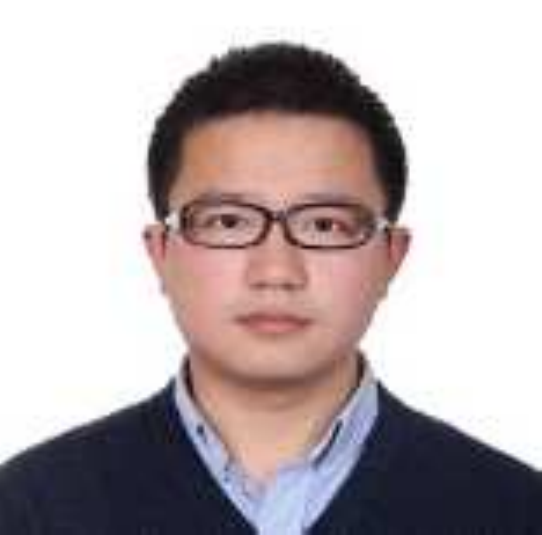}}]{Liang Chen}
  received the bachelor’s and Ph.D. degrees from Zhejiang University (ZJU) in 2009 and 2015, respectively. He is currently an associate professor with the School of Computer Science and Engineering, Sun Yat-Sen University (SYSU), China. His research areas include data mining, graph neural network, adversarial learning, and services computing. In the recent five years, he has published over 70 papers in sev- eral top conferences/journals, including SIGIR, KDD, ICDE, WWW, ICML, IJCAI, ICSOC, WSDM, TKDE, TSC, TOIT, and TII. His work on service recommendation has received the Best Paper Award Nomination in ICSOC 2016. Moreover, he has served as PC member of several top conferences including SIGIR, WWW, IJCAI, WSDM etc., and the regular reviewer for journals including TKDE, TNNLS, TSC, etc.
\end{IEEEbiography}

\begin{IEEEbiography}[{\includegraphics[width=1in,height=1.25in,clip,keepaspectratio]{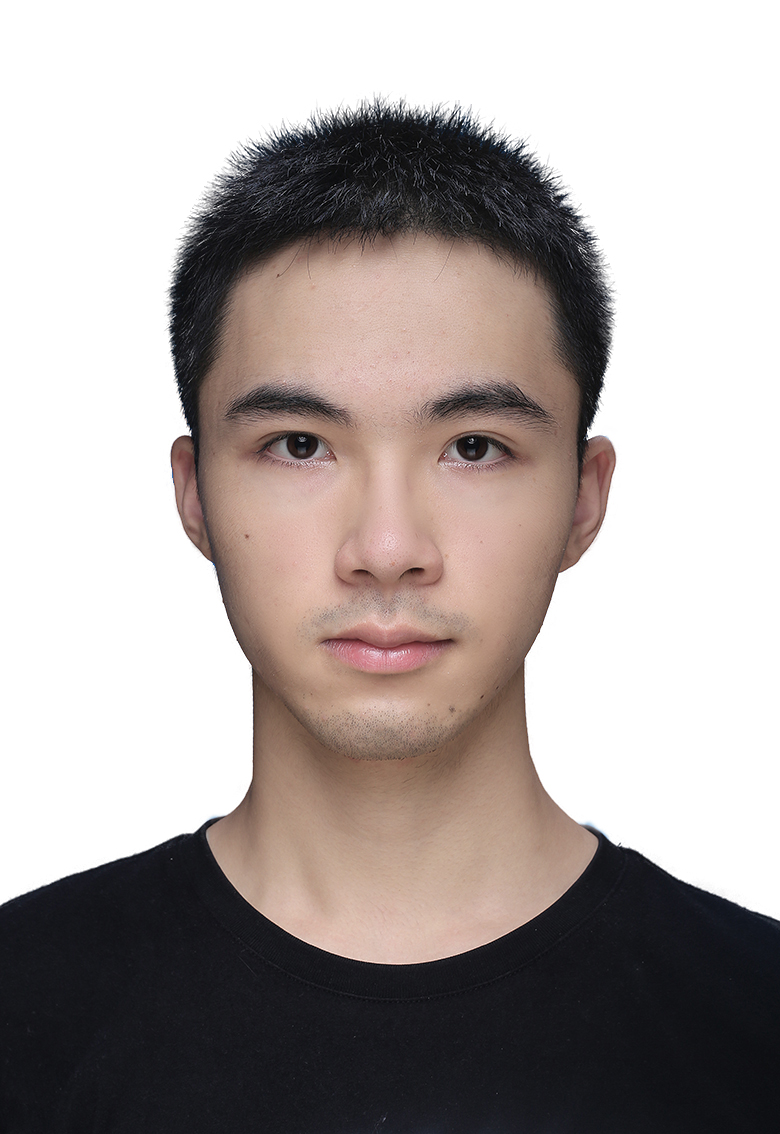}}]{Qibiao Peng}
 received the bachelor's degree at Chongqing University (CQU) in 2020. He is currently pursuing the master's degree with the School of Computer Science and Engineering, Sun Yat-sen University, Guangzhou, China. His main research interests include graph representation learning, machine learning, and code representation learning.
\end{IEEEbiography}

\begin{IEEEbiography}[{\includegraphics[width=1in,height=1.25in,clip,keepaspectratio]{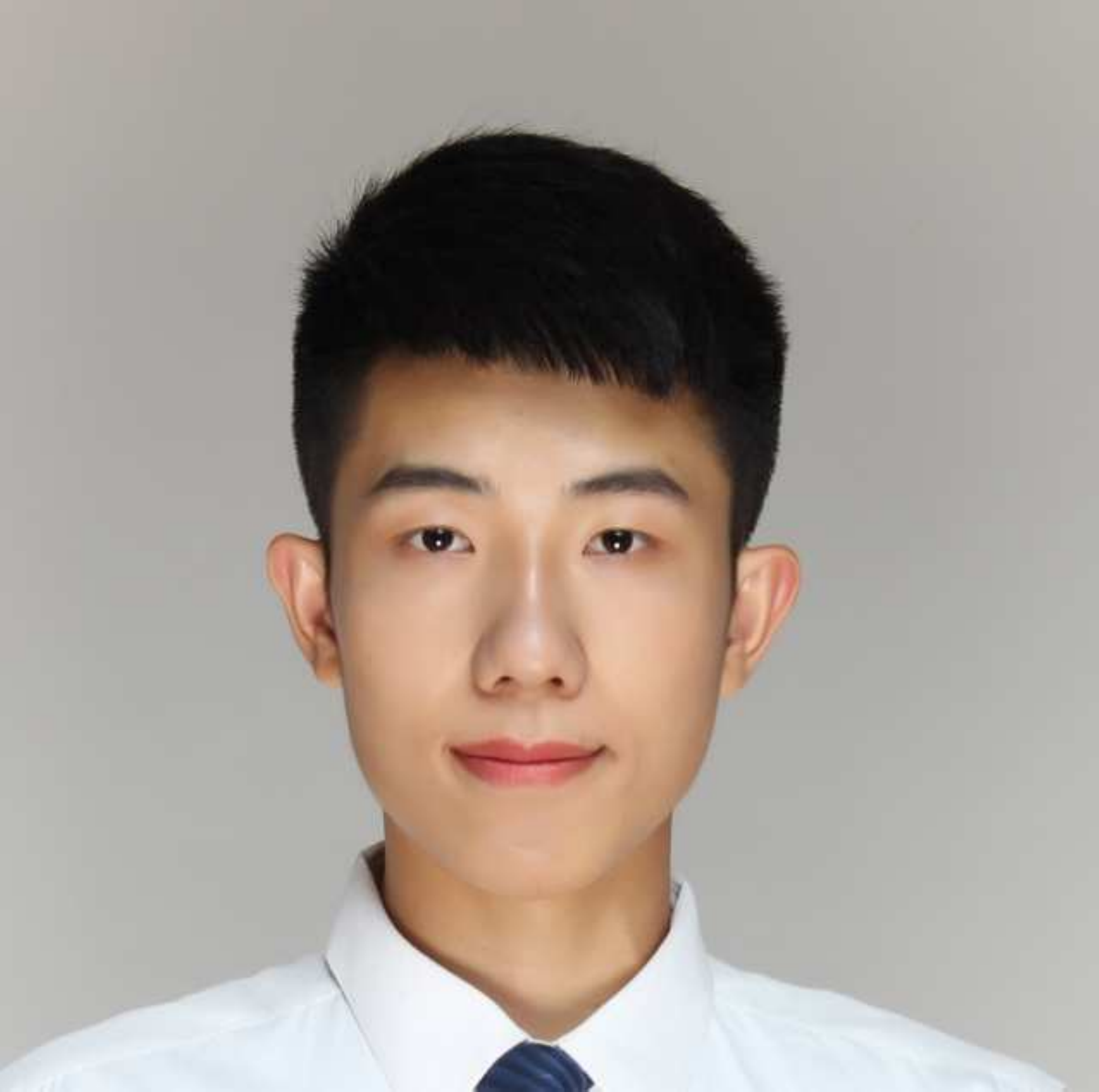}}]{Jintang Li}
  received the bachelor's degree at Guangzhou Medical University, Guangzhou, China, in 2018. He is currently pursuing the master's degree with the School of Electronics and Communication Engineering, Sun Yat-sen University, Guangzhou, China. His main research interests include graph representation learning, adversarial machine learning, and data mining techniques.
\end{IEEEbiography}

\begin{IEEEbiography}[{\includegraphics[width=1in,height=1.25in,clip,keepaspectratio]{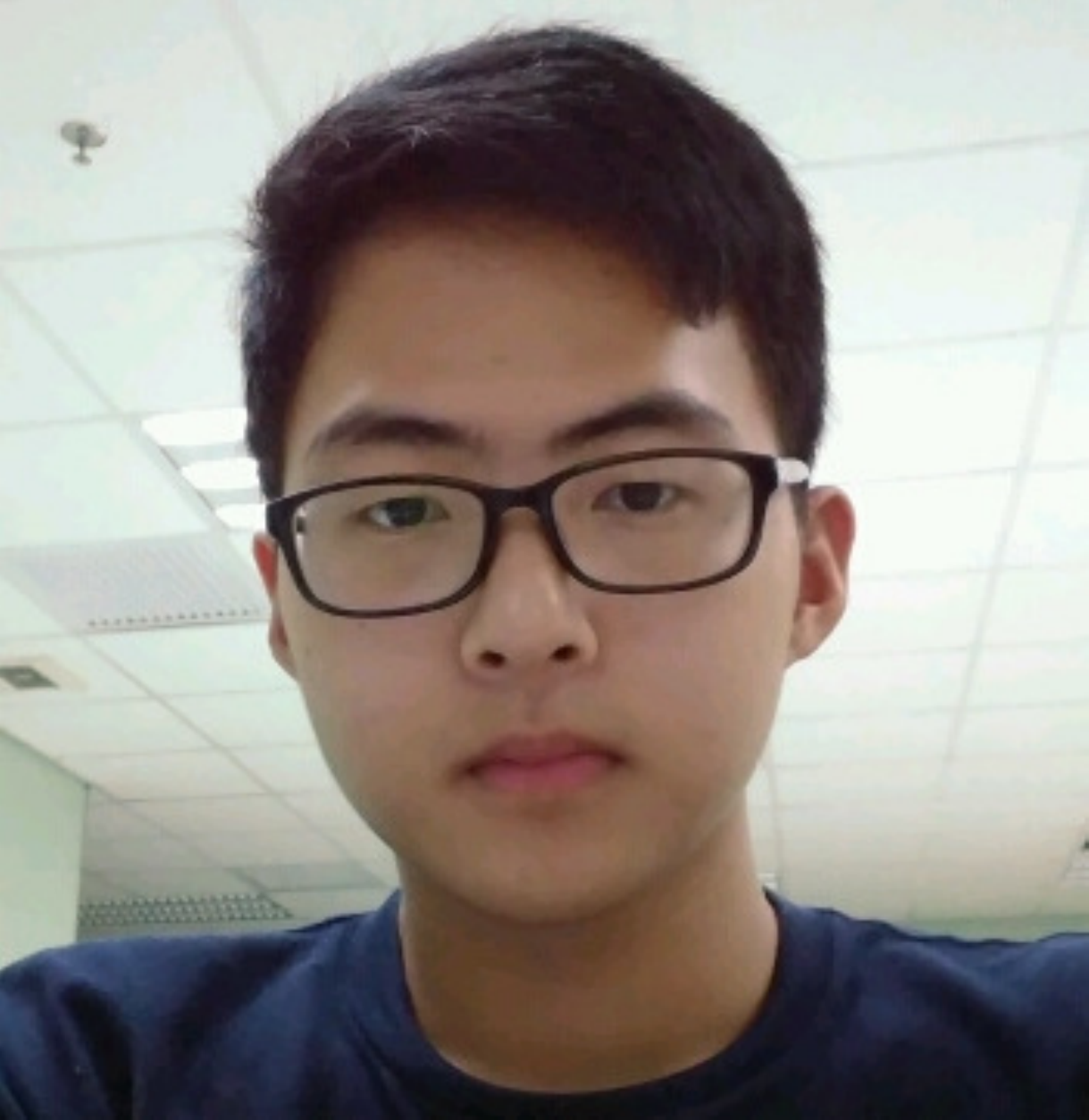}}]{Yang Liu}
  received the bachelor’s degree at Sun Yat-sen University, Guangzhou, China, in 2019. He is currently pursuing the master’s degree with the School of Data and Computer Science, Sun Yat-sen University, Guangzhou, China. His main research interests include recommendation systems, machine learning and data mining techniques.

\end{IEEEbiography}

\begin{IEEEbiography}[{\includegraphics[width=1in,height=1.25in,clip,keepaspectratio]{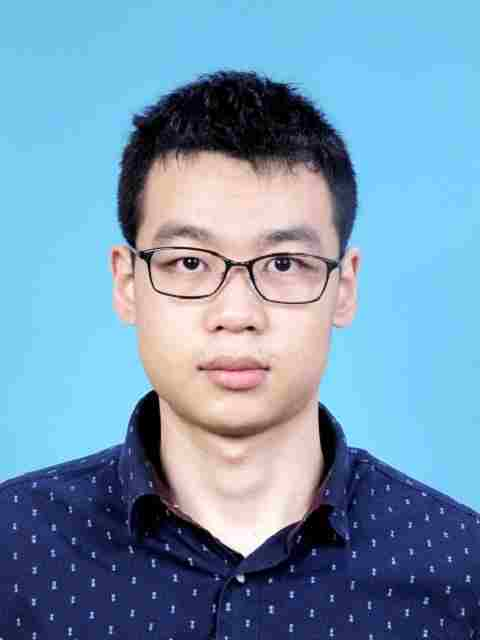}}]{Jiawei Chen}
  is currently a Posdoc Research Fellow in LDS (Lab of Data Science), School of Information Science and Technology, University of Science and Technology of China. His research interests include information retrieval, data mining and machine learning, particularly in recommender systems, sampling, graph neural network. Moreover, He have served as the PC member for top-tier conferences including SIGIR, ACMMM, PKDD-ECML and the invited reviewer for prestigious journals such as TNNLS, TKDE, TOIS.
\end{IEEEbiography}

\begin{IEEEbiography}[{\includegraphics[width=1in,height=1.25in,clip,keepaspectratio]{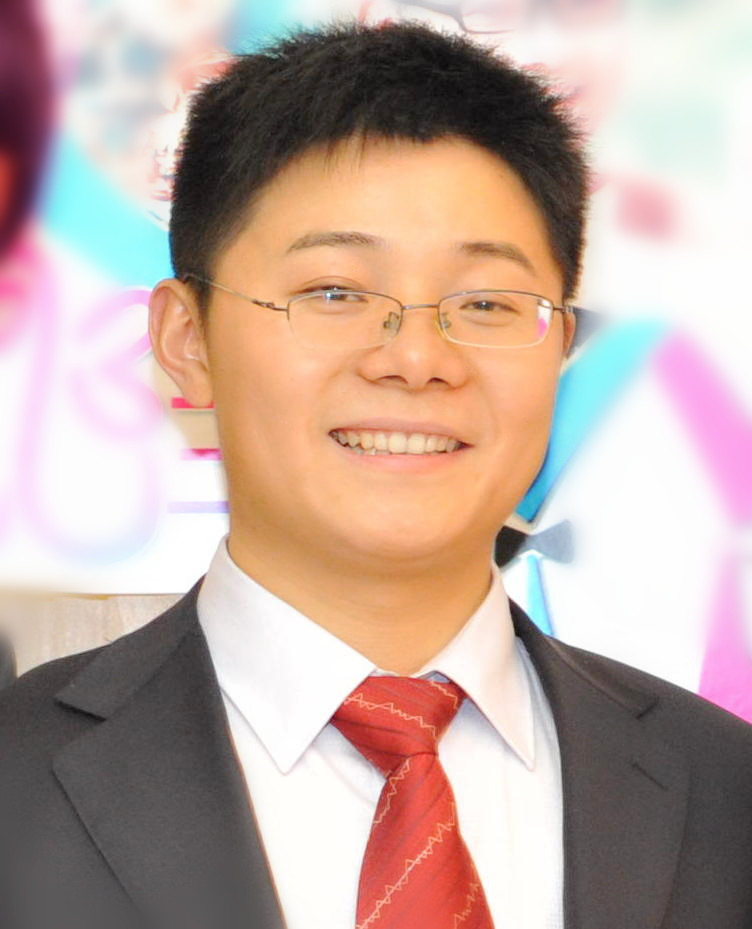}}]{Yong Li}
  received the B.S. degree from Huazhong University of Science and Technology in 2007, and the M. S. and the Ph. D. degrees in Electrical Engineering from Tsinghua University, in 2009 and 2012, respectively. During 2012 and 2013, he was a Visiting Research Associate with Telekom Innovation Laboratories and Hong Kong University of Science and Technology respectively. During 2013 to 2014, he was a Visiting Scientist with the University of Miami. Currently, he is a Faculty Member of the Department of Electronic Engineering, Tsinghua University. His research interests are in the areas of big data, mobile computing, wireless communications and networking.
\end{IEEEbiography}

\begin{IEEEbiography}[{\includegraphics[width=1in,height=1.25in,clip,keepaspectratio]{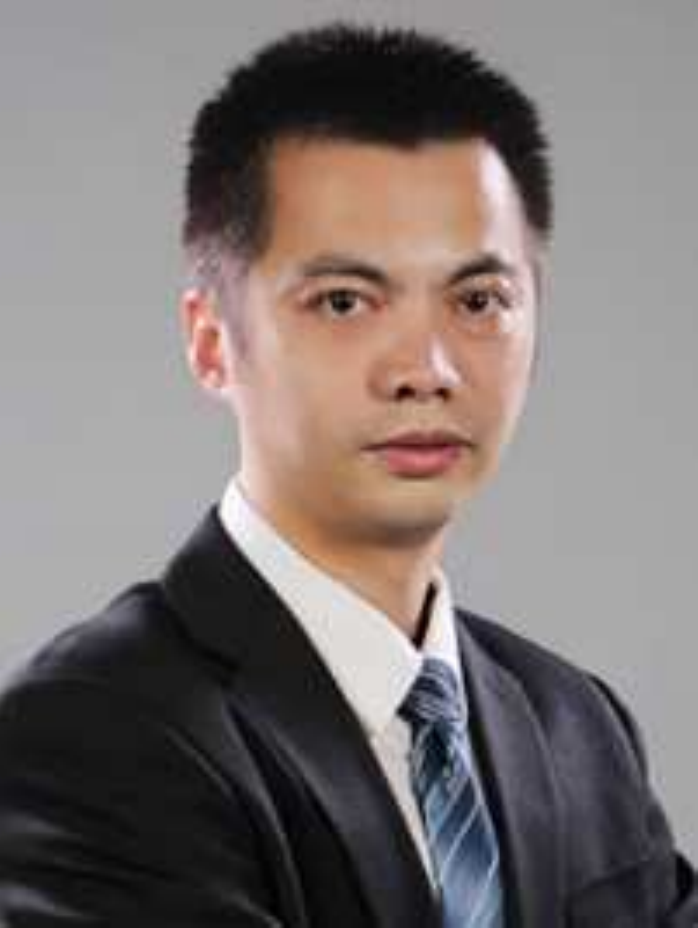}}]{Zibin Zheng}
  received the Ph.D. degree from The Chinese University of Hong Kong, in 2011. He is currently a Professor with the School of Computer Science and Engineering, Sun Yat-sen University, Guangzhou, China. His research interests include services computing, software engineering, and blockchain. He received the ACM SIGSOFT Distinguished Paper Award at the ICSE’10, the Best Student Paper Award at the ICWS’10, and the IBM Ph.D. Fellowship Award.
\end{IEEEbiography}


\end{document}